\documentclass[journal]{IEEEtran}
\IEEEoverridecommandlockouts
\usepackage{color}
\usepackage{tikz}
\usetikzlibrary{shapes, arrows, arrows.meta}
\usepackage{mathtools}
\usepackage{amssymb}
\usepackage{lipsum}
\usepackage{float}
\usepackage{makecell}
\usepackage{multirow}
\usepackage{longtable}
\usepackage{stfloats}
\usepackage{pifont}
\usepackage{xcolor}

\usepackage{multirow}
\usepackage{amsmath}
\usepackage{bm}
\usepackage{algorithm}
\usepackage{algpseudocode}
\usepackage{booktabs}
\usepackage[numbers,sort&compress]{natbib}
\usepackage{balance}
\usepackage{soul}
\algdef{SE}[SUBALG]{Indent}{EndIndent}{}{\algorithmicend\ }%
\algtext*{Indent}
\algtext*{EndIndent}
\usepackage{xcolor}
\usepackage{threeparttable}
\usepackage{natbib}
\usepackage[citecolor=blue,urlcolor=blue,bookmarks=false,hypertexnames=true]{hyperref} 

\usepackage{graphicx}
\usepackage{caption}
\usepackage{subfigure}
\begin{document}

% 自定义一个新的命令来改变标题的颜色

\title{LEGO: Learning and Graph-Optimized Modular Tracker for Online Multi-Object Tracking with Point Clouds}

\author{\IEEEauthorblockA{
Zhenrong Zhang\IEEEauthorrefmark{1},
Jianan Liu\IEEEauthorrefmark{1},
Yuxuan Xia,
Tao Huang,~\IEEEmembership{Senior Member,~IEEE,}\\
Qing-Long Han,~\IEEEmembership{Fellow,~IEEE},
and 
Hongbin Liu\IEEEauthorrefmark{2},~\IEEEmembership{Member,~IEEE}
}
\vspace{-5 mm}

\thanks{This work has been submitted to the IEEE for possible publication. Copyright may be transferred without notice, after which this version may no longer be accessible.}
\thanks{\IEEEauthorrefmark{1}Both authors contribute equally to the work and are co-first authors.}
\thanks{\IEEEauthorrefmark{2}Corresponding author.}
\thanks{Z.~Zhang is with the School of AI and Advanced Computing, Xi'an Jiaotong-Liverpool University, Suzhou, P.R.~China. Email: Zhenrong.Zhang21@student.xjtlu.edu.cn.}
\thanks{J.~Liu is with Momoni AI, Gothenburg, Sweden. Email: jianan.liu@momoniai.org.}
\thanks{Y. Xia is with the Department of Automation, Shanghai Jiaotong University, Shanghai 200240, China. Email: yuxuan.xia@sjtu.edu.cn.}
\thanks{T.~Huang is with College of Science and Engineering, James Cook University, Cairns, QLD 4878, Australia. Email: tao.huang1@jcu.edu.au.}
\thanks{Q.-L.~Han is with the School of Engineering, Swinburne University of Technology, Melbourne, VIC 3122, Australia. Email: qhan@swin.edu.au.}
\thanks{H.~Liu is with the School of AI and Advanced Computing, Xi'an Jiaotong-Liverpool University, Suzhou, P.R.~China. Email: Hongbin.Liu@xjtlu.edu.cn.}
}

%\markboth{IEEE Transactions on Intelligent Transportation Systems, Submitted in Aug.~2024}
%{\MakeLowercase{\textit{et al.}}: Demo of IEEEtran.cls for IEEE Journals}

\maketitle

\begin{abstract}

Online Multi-Object Tracking (MOT) plays a pivotal role in autonomous systems. The state-of-the-art approaches usually employ a tracking-by-detection method, and data association plays a critical role. This paper proposes a learning and graph-optimized (LEGO) modular tracker to improve data association performance in the existing literature. The proposed LEGO tracker integrates graph optimization, which efficiently formulates the association score map, facilitating the accurate and efficient matching of objects across time frames. To further enhance the state update process, the Kalman filter is added to ensure consistent tracking by incorporating temporal coherence in the object states to further enhance the state update process. Our proposed method, utilising LiDAR alone, has shown exceptional performance compared to other online tracking approaches, including LiDAR-based and LiDAR-camera fusion-based methods. LEGO ranked $3^{rd}$ among all trackers (both online and offline) and $2^{nd}$ among all online trackers in the KITTI MOT benchmark for cars\footnote{\href{https://www.cvlibs.net/datasets/kitti/eval_tracking.php}{https://www.cvlibs.net/datasets/kitti/eval\_tracking.php}}, at the time of submitting results to KITTI object tracking evaluation ranking board. Moreover, our method also achieves competitive performance on the Waymo open dataset benchmark.

\end{abstract}

% Note that keywords are not normally used for peer-review papers.

%Too many keywords. 3-5 keywords are quite enough.        Commented by BZ

\IEEEpeerreviewmaketitle
\setcitestyle{square}

\section{Introduction}
\label{introduction}

\IEEEPARstart Tracking is a crucial technology utilized in various systems within the field of intelligent transportation systems (ITS), like pedestrian tracking \cite{Pedestrian_MOT, Pedestrian_MOT_2}, drone monitoring \cite{Drone_ITS_MOT, Drone_ITS_MOT_2}, target tracking \cite{3dsot}, traffic monitoring \cite{MOT_ITS, ITS_TPMBM}, Advanced Driver Assistance Systems (ADAS) and Autonomous Driving (AD) \cite{GM-PHD_MOT, CAMO-MOT}. To develop robust and accurate tracking systems effectively, various approaches have been investigated by using different sensor modalities, encompassing 2D multi-ojbect-tracking (MOT) methods using camera images, 3D MOT techniques relying solely on LiDAR, and 3D MOT methods that leverage both camera and LiDAR. In the realm of 2D MOT, notable methods, such as FairMOT \cite{zhang2021fairmot}, ByteTrack \cite{zhang2021bytetrack}, and StrongSort \cite{Strongsort} predominantly employ camera data for tracking objects. Although these methods provide valuable insights, their performance is constrained by the limitations of working with 2D representations alone. On the other hand, 3D MOT techniques exclusively employ LiDAR to capture 3D information about the tracked objects. AB3DMOT \cite{AB3DMOT} serves as an exemplary method within this category, demonstrating the efficacy of LiDAR-based approaches for precise object localization and spatial understanding. Furthermore, there are 3D MOT methods that exploit the synergies between camera and LiDAR systems. Probabilistic CBMOT \cite{CBMOT} represents one such approach, where both sensor modalities are utilized to improve tracking performance. By combining the strengths of cameras and LiDAR sensors, these methods aim to achieve enhanced object tracking accuracy. A majority of these MOT methods can be classified as tracking-by-detection methods, in which various 3D object detectors, e.g., PointRCNN \cite{PointRCNN}, PointGNN \cite{PointGNN}, CenterPoint \cite{CenterPoint}, and CasA \cite{CasA}, etc, have been employed to provide estimated bounding boxes as measurements input to the MOT pipeline. Some work, such as \cite{Drone_ITS_MOT, Drone_ITS_MOT_2}, may use drone monitoring to collect 2D video image information and perform the tracking based on a temporal transformer.  As drone video is usually two-dimensional, it lacks the rich depth information available from other sources. However, continuous point cloud data is a form of 3D video for real-world representation. This is quite different from conventional 2D video, as it captures spatial structure and depth information directly in three dimensions, enabling more accurate modeling, understanding, and tracking of dynamic scenes in complex environments.

Despite many efforts have been made in this area, MOT systems still suffer from erroneous measurements generated from object detector and inaccurate data association. Specifically, erroneous measurements from object detectors lead to misaligned or imprecise bounding boxes, which degrade the quality of input data for tracking. Inaccurate data association, on the other hand, results in track identity switches or fragmented tracks, making it difficult to maintain consistent object trajectories. Consequently, these issues significantly hinder the overall performance of MOT systems.

To address these challenges, we propose a novel framework that explicitly tackles both issues through the introduction of two key components: the offset correction module and the Adjacency Matrix guided Graph Network (AMGN) score calculation module. The offset correction module is designed to correct the erroneous bounding boxes generated by the detector, resulting in more accurate and reliable object localization. Meanwhile, the AMGN score calculation module redefines the data association problem as a bipartite graph matching problem, leveraging a combination of graph structure information, an LSTM, and a new self-attention structure whose coefficients is derived from the learnable adjacency matrix to enhance relational reasoning and reduce information redundancy. This integration allows the Graph Neural Network (GNN) to effectively capture complex dependencies between measurements and predicted tracks, improving the accuracy of data association. Lastly, by integrating with Kalman filter, our approach facilitates more robust state updates and improves overall tracking performance.

The contributions of this paper are outlined as follows:
\begin{itemize}

\item{
A easily comprehensible and tunable LEarning and Graph-Optimized (LEGO) modular tracker is proposed for online MOT. LEGO introduces two learning modules as its core, an offset correction module and a Adjacency Matrix guided Graph Network (AMGN) score calculation module.
}
\item{
The offset correction module is designed to predict the offset between these detection results and the ground truth. The primary purpose of this mechanism is to rectify prediction outcomes, thereby improving the accuracy of the model's predictive capabilities. The AMGN score calculation module which contains LSTM and $K$ AMGN blocks, leverages the LSTM and graph structure information to perform message passing and update node features using the self-attention coefficients derived from a learnable adjacency matrix. By doing so, the AMGN score calculation module can directly computes the association score.
}
\item{
Empirical evaluation and performance analysis of the proposed offset correction and AMGN score calculation modules, demonstrate their effectiveness in improving tracking performance, as well as the proposed LEGO tracker's state-of-the-art performance in both the KITTI MOT benchmark and Waymo 3D MOT benchmark.
}
\end{itemize}

The structure of this paper is organized as follows: Section \ref{relatedwork} presents a thorough review of related works in the field of LiDAR-based MOT within the context of autonomous driving applications. Section \ref{LEGO_Approach} introduces the proposed LEGO Modular tracker, elucidating its key components and functionalities. Subsequently, in Section \ref{experiment}, the experimental results are presented and analyzed in detail. Finally, the paper concludes with Section \ref{conclusion}, summarizing the key findings and contributions of the study.

%==============================================================
\section{Related Work}\label{relatedwork}

%--------------------------------------------------
\subsection{MOT with LiDAR Only}
\subsubsection{Model-based Methods}
Similar to MOT tasks in image filed \cite{newcrf}\cite{sqgrl}, contemporary tracking systems in ADAS and AD applications often employ global nearest neighbor methods and heuristics for data association. In LiDAR-based MOT, several methods have emerged that rely exclusively on LiDAR sensors. Chiu et al.  \cite{StanfordIPRL-TRI} made a pioneering contribution by integrating Mahalanobis distance with AB3DMOT, establishing a benchmark for addressing LiDAR-based 3D MOT challenges. Similarly, SimpleTrack \cite{SimpleTrack} introduced a generalized version of 3D IoU, known as GIoU, as the scoring mechanism for tracking-by-detection tasks. Bytetrackv2 \cite{bytetrackv2} employed a hierarchical data association strategy to identify genuine objects within low-score detection boxes, effectively mitigating issues related to object loss and fragmented trajectories. Additionally, this system employed Non-Maximum Suppression (NMS) to preprocess object detections. Empirical evidence demonstrated that combining GIoU with NMS preprocessing enhanced overall tracking performance.

Maintaining tracks even when objects are no longer visible is a common feature shared by ImmortalTracker \cite{ImmortalTracker} and PC3T \cite{PC3T}. This helps reduce identification switches and fragmented tracks. ACK3DMOT \cite{ACK3DMOT} introduced a cost matrix for tracking-by-detection tasks based on a joint probability function that considers appearance, geometry, and distance correlation between detected bounding boxes and predicted objects. When combined with an adaptive cubature Kalman filter, this approach achieved enhanced tracking performance. PF-MOT \cite{PF-MOT} utilized a cluster-based earth-mover distance, Euclidean distance, and feature similarity to construct the cost matrix. Considering uncertainties, UG3DMOT \cite{UG3DMOT} evaluated data association based on random vectors, where the similarity between two multidimensional distributions was evaluated using the Jensen-Shannon divergence.

As the raise of applying random finite set (RFS) in MOT applications \cite{GM-PHD_MOT}, there are alternative algorithms where each potential object is modelled using a Bernoulli process with probabilistic object existence. RFS-M3 \cite{Pang2021PMBM} utilized the Poisson multi-Bernoulli mixture filter based on a random finite set (RFS) to tackle the LiDAR-based MOT problem. Through systematic comparative analysis, GNN-PMB \cite{GNN-PMB} demonstrated that the contemporary RFS-based Bayesian tracking framework outperformed the traditional random vector-based Bayesian tracking framework. BP-Tracker \cite{BP_Tracker} presented a factor graph formulation of the MOT problem and employed a belief propagation algorithm to compute the marginal association probability, representing a significant advancement in the field.

\subsubsection{Deep-learning based Methods}
SimTrack \cite{SimTrack} and CenterTube \cite{CenterTube} introduce end-to-end trainable models for joint detection and tracking, leveraging raw point cloud as input. OGR3MOT \cite{OGR3MOT}, Batch3DMOT \cite{Batch3DMOT}, PolarMOT \cite{PolarMOT}, Rethinking3DMOT propose sophisticated graph structures based on neural message passing, enabling online execution of detection and tracking processes. The ENBP-Tracker \cite{ENBP_Tracker} integrates GNN into its design, distinguishing itself by combining the network with a belief propagation tracker. This hybrid tracking architecture enhances the robustness and efficacy of the tracking process, showcasing the potential of merging traditional tracking methods with advanced neural network architectures. Intertrack \cite{Intertrack} TransMOT \cite{TransMOT} adopt transformer structures to generate discriminative object representations for data association. Minkowski-Tracker \cite{Minkowski_Tracker}, PC-TCNN \cite{PC-TCNN} and ShaSTA \cite{ShaSTA}, employ proposal networks to extract features from various feature maps, facilitating the learning of affinity matrix for point cloud-based MOT.

\subsection{MOT with LiDAR and Camera Fusion}
The effectiveness of MOT can be further enhanced by adopting a fusion approach that combines data from LiDAR and camera sources. Several methods, such as Probabilistic3DMM \cite{Probabilistic3DMM}, CBMOT \cite{CBMOT}, GNN3DMOT \cite{Gnn3dmot}, and MF-Net \cite{MF-Net} have utilized this fusion approach to leverage the complementary strengths of different sensor modalities, leading to more comprehensive and precise object tracking. IMSF MOT \cite{IMSF-MOT} proposes a novel feature fusion method using Pointnet++ to extract more discriminative features and improve multiple object tracking performance.

DeepFusionMOT \cite{DeepFusionMOT} and EagerMOT \cite{EagerMOT} improve tracking performance compared to Probabilistic3DMM by implementing a two-stage data association scheme. This scheme leverages 3D detection data obtained from LiDAR and camera inputs, and 2D detection data obtained solely from the camera. By combining information from multiple modalities, a more comprehensive understanding of the environment is achieved, potentially enhancing the robustness and accuracy of the tracking process. Building upon EagerMOT, AlphaTrack \cite{AlphaTrack} introduces a feature extractor that concatenates image and point cloud information as input to enhance performance. Additionally, other relevant works, such as DualTracker \cite{DualTracker}, HIDMOT \cite{HIDMOT}, CAMO-MOT \cite{CAMO-MOT}, MSA-MOT \cite{MSA-MOT}, and JMODT \cite{JMODT} utilize PointGNN or PointRCNN as their 3D object detectors to estimate object detections. They are combined with detection proposals from image data, and a hybrid multi-modal input is employed for the association within the tracking-by-detection task, potentially improving the overall performance and accuracy of the tracking system.

\subsection{MOT with Graph Neural Network based Methods}
In recent years, GNN-based tracking methods have attracted considerable research interest, primarily due to their capability to model Multi-Object Tracking (MOT) as a graph structure, allowing object nodes to effectively exchange and aggregate features. GNN3DMOT \cite{Gnn3dmot} was among the first to propose a GNN-based 3D MOT framework, integrating both 2D and 3D feature learning. Their method employs a feature extractor to obtain motion and appearance features from both 2D images and 3D point clouds, subsequently feeding these features into a GNN to facilitate feature interaction among objects.
Furthermore, \cite{GNNMOT} introduced an MOT approach comprising two separate graph networks: an appearance graph network and a motion graph network. These two GNNs independently compute similarity measures between detected objects and existing trackers through a four-step updating module. The final affinity matrix is then derived by combining these two similarity scores through a weighted fusion mechanism.
Additionally, \cite{MOTG} proposed an iterative Graph Convolutional Network (GCN) clustering method designed to reduce computational complexity. This method ranks generated proposals based on estimated quality scores, effectively maintaining proposal quality while significantly enhancing computational efficiency. Similarly, \cite{MOTG2} presented a GCN-based MOT approach, modeling the relationships between existing tracklets and intra-frame detections as a general undirected graph to improve tracking robustness and accuracy. Most existing methods perform message passing and updating with Graph Attention Networks (GAT) to ensure that different nodes contribute differently during the message passing process. However, GAT handles message passing and node updating on a node-by-node basis, which overlooks the global structure of the graph. This limitation may result in the loss of critical information embedded within the overall graph structure. To address this issue, we propose the AMGN block, which incorporates a learnable adjacency matrix into the self-attention mechanism during the message passing and updating process. This design enables the model to capture both local and global structural information, leading to more effective graph representation.

\section{The Proposed LEGO Modular Tracker}
\label{LEGO_Approach}

\begin{figure*}[!t]
\centering
\includegraphics[width=\linewidth]{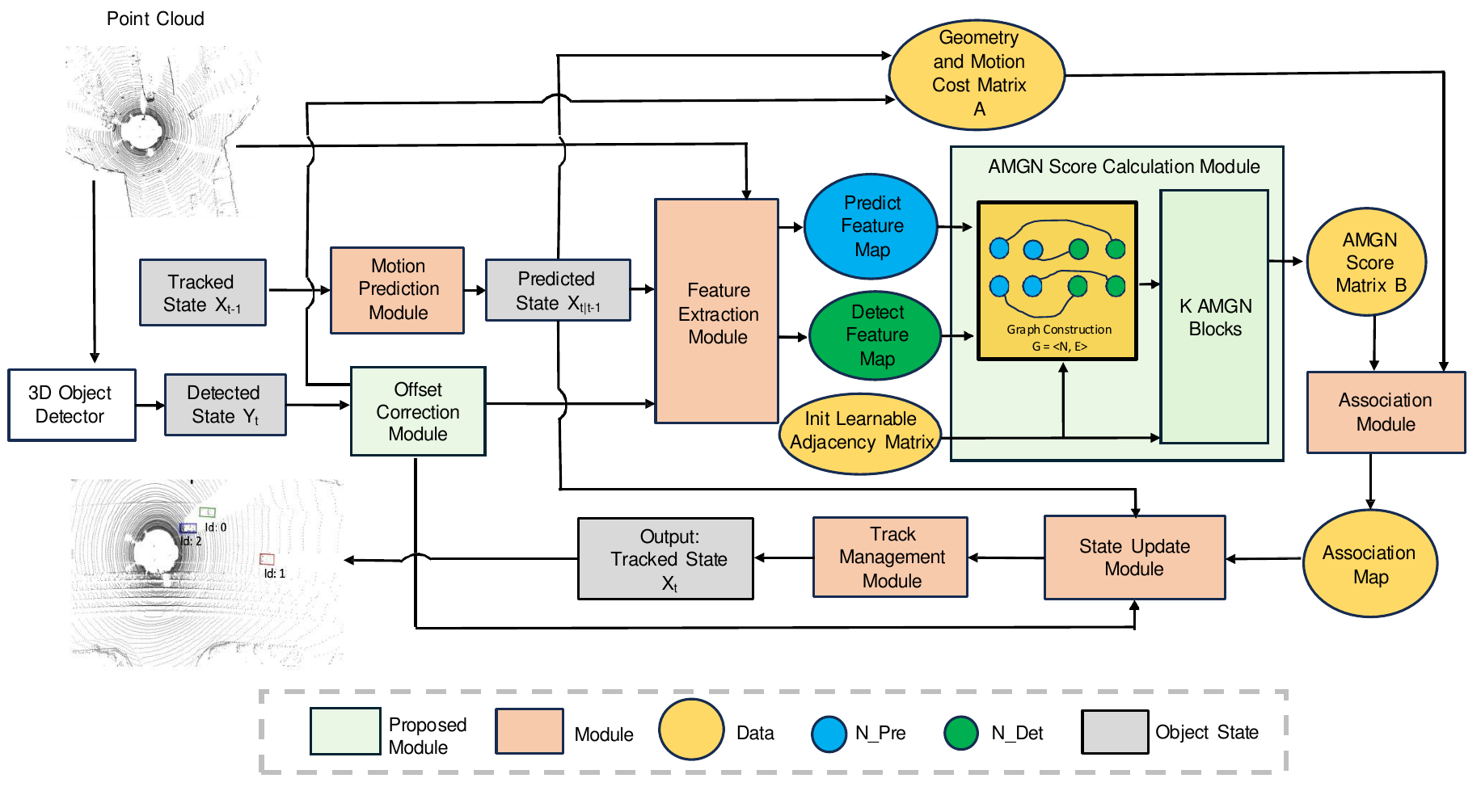}

\caption{The proposed LEGO tracker seamlessly integrates several modules. First, the offset correction module refines detection results and corrects detected errors. Simultaneously, the motion prediction module utilizes the tracked state $\bm{X}_{t}$ to forecast the next state $\bm{X}_{t|t-1}$. The feature extraction module then derives 3D feature maps for both predicted and detected objects. Using this information, the AMGN score calculation module constructs a bipartite graph and computes the AMGN score matrix. The association module subsequently integrates the geometry and motion cost matrix with the AMGN score matrix, yielding an association map. Finally, the state update module refines the predicted state, and the track management module processes the matched detected states, completing the tracking cycle.}
\label{framework}
\end{figure*}

Within this section, we will introduce the LEGO modular tracker. Firstly, we will present its overall framework, followed by a detailed elaboration of each module. Fig. \ref{framework} shows the proposed LEGO tracker, which comprises several modules: motion prediction, proposed offset correction, feature extraction, proposed AMGN score calculation, association, state update, and track management.

\subsection{The Overview Framework of LEGO Modular Tracker} 
Initiating the process, the 3D point cloud is fed into a 3D object detector. To refine these detection results, an offset correction module is introduced. This module predicts the offset between the detection outputs and the ground truth, smoothing the detector's predictions. Simultaneously, a motion prediction module uses the tracked state from the previous time step to predict the current state of each object, by using Constant Acceleration (CA) motion model \cite{bayes}.

Next, the outputs from the motion prediction and offset correction modules, along with the 3D point cloud, are passed through a feature extraction module. This module, based on the PointNet architecture \cite{qi2017pointnet}, generates robust 3D feature maps. Using these feature maps, a bipartite graph $\bm{G}=\langle \bm{N},\bm{E} \rangle$ is constructed, where $\bm{N}$ represents nodes for detected and predicted objects, and $\bm{E}$ represents potential associations between them. The bipartite graph $\bm{G}$ and a learnable adjacency matrix are then input into the AMGN block under the proposed AMGN score calculation module, which outputs the AMGN score matrix $\bm{B}$. Using LSTM and $K$ AMGN blocks with the learnable adjacency matrix, the AMGN score calculation module captures and refines the structural relationships within the graph, quantifying the relational strengths between nodes. The final association cost matrix is derived by combining the AMGN score matrix $\bm{B}$, with the geometry and motion cost matrix $\bm{A}$ which is calculated by following PC3T \cite{PC3T}.

The data association module then matches detected and predicted objects based on the association cost matrix. Finally, the resulting association map is passed to the state update module, which updates object states using the standard Kalman filter update. The track management module handles unmatched detected states and predicted states by using classical $M/N$ logic. Specifically, if a predicted state remains unmatched for more than $N_{t}$ consecutive frames, it is removed as leaving the scene. Otherwise, unmatched predicted states are retained, assuming the object may reappear in subsequent frames. The track management module also consider all unmatched objects as potential new objects entering the scene. In order to avoid creating false positive trajectories, new trajectory will not be created for the unmatched detection until it has been continually matched in the next $M_{t}$ frames.

\vspace{-5pt}
%% review1 q2
\subsection{Motion Prediction Module}
Certain MOT techniques prefer to use a constant-velocity motion model based on the Kalman Filter. This model assumes that the object moves at a nearly constant speed, which may not hold true in several real-world situations. As a result, consecutive missed detections can lead to considerable errors in motion prediction. To overcome this problem, the constant acceleration (CA) motion model \cite{bayes} is used based on the Kalman Filter, which offers a more precise representation of the object state. In the CA motion model, the object state $\bm{X}_{t-1}$ at a specific temporal instance $t-1$ is represented as $ \left[\bm{pos}_{t-1}, \bm{v}_{t-1}, \bm{\alpha}_{t-1}\right]^T$, where $\bm{pos}_{t-1}$ denotes the position, $\bm{v}_{t-1}$ represents the velocity, and $\bm{\alpha}_{t-1}$ indicates the orientation of the object. The prediction of the mean state $\bm{X}_{t | t-1}$ and covariance $\bm{P}_{t | t-1}$ using Kalman prediction equations with the CA motion model is as follows:
\begin{equation}
\label{1}
\bm{X}_{t|t-1} = \bm{S}_{A}\bm{X}_{t-1},
\end{equation}
\begin{equation}
\label{2}
\bm{P}_{t|t-1} = \bm{S}_{A}\bm{P}_{t-1}\bm{S}_{A}^{T}+\bm{Q},
\end{equation}
where $\bm{S}_{A}$ is the state transition matrix, $\bm{Q}$ represents the motion noise covariance matrix, $\bm{I}$ denotes the identity matrix, $\bm{O}$ represents the zero matrix, $n$ denotes the state dimension, and $\delta$ and $\mathbf{a}$ are sensor-related hyper-parameters.

\begin{figure}[t]
\centering 
\includegraphics[width=0.48\textwidth]{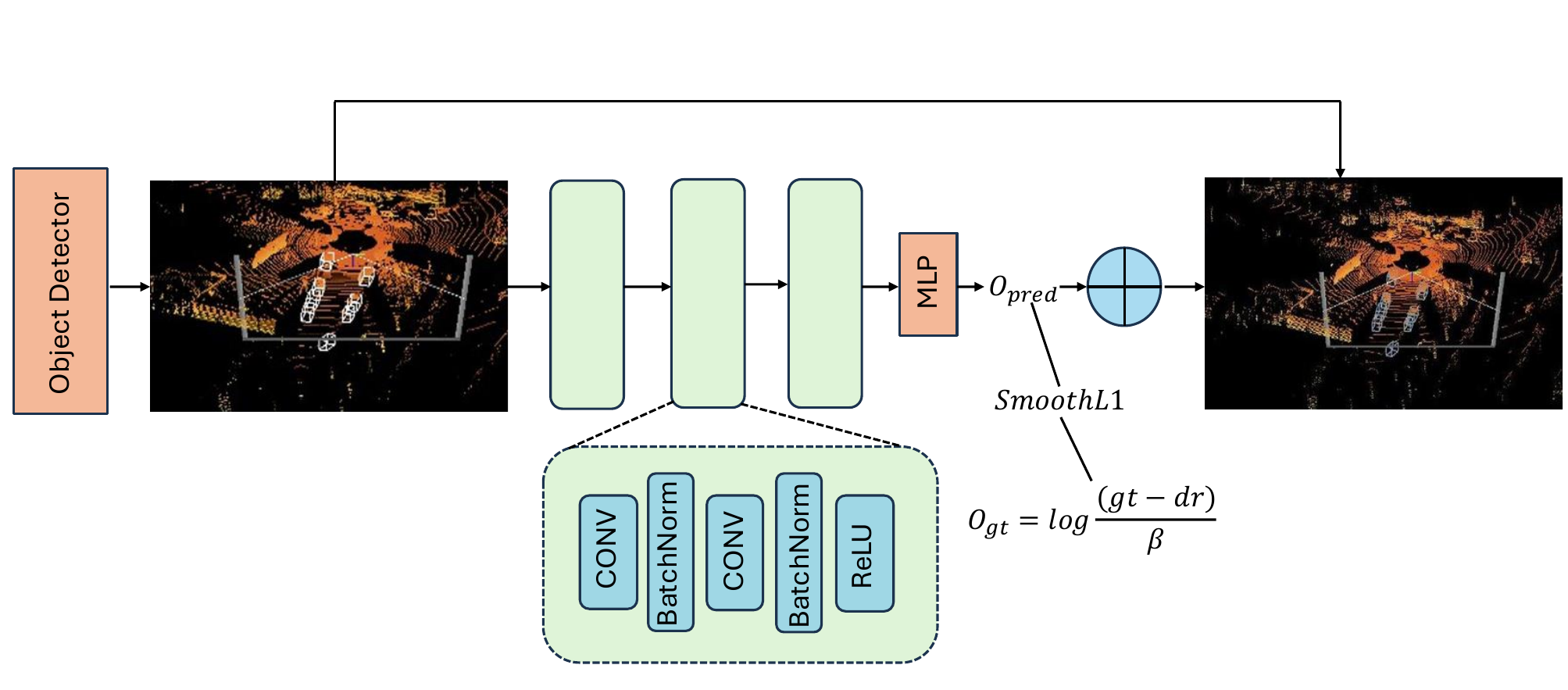} 
\caption{The structure of the offset correction module. The detection result, which comprises the parameters $x_{p}^{d}, y_{p}^{d}, z_{p}^{d}, w^{d}, h^{d}, l^{d}$, is input into the offset prediction head. The offset prediction head can be constructed" as a simple neural network, which computes the offset between the ground truth and the obtained detection result.}
\label{dethead} 
\end{figure}

\subsection{Offset Correction Module}

Our investigation has revealed a significant issue related to the accuracy of the state inferred by the detector, especially when errors arise from the 3D object detector. In order to mitigate this potential source of inaccuracies, an offset correction module is proposed. This module serves the crucial purpose of rectifying the detection outcomes generated by the 3D object detector, thereby enhancing the overall accuracy of the system. The architecture and components of this offset correction module are depicted in Fig. \ref{dethead}.
This module comprises a 3D convolution layer, batch normalization layer, and Multi-Layer Perception (MLP) with the residual connection. The offset obtained from the head is $O_{pred}$, and the ground truth offset is:
\begin{equation}
\label{offset}
O_{gt} = \log\left(\frac{gt-dr}{\beta}\right),
\end{equation}
where ${gt=\left(x_{p}^{g},y_{p}^{g},z_{p}^{g},w^{g},h^{g},l^{g}\right)}$ represents the ground truth bounding box coordinates  $(x_p^g,y_p^g,z_p^g)$, width $w^g$, height $h^g$, length $l^g$; ${dr=\left(x_{p}^{d},y_{p}^{d},z_{p}^{d},w^{d},h^{d},l^{d}\right)}$ indicates the object detection bounding box extracted from the 3D object detector, and $\beta$ is the scaling factor. 

The offset correction module is directly supervised by the ground-truth offset. The Hungarian algorithm is used to match the ground truth bounding box and the predicted bounding box. Specifically, the loss function is defined for training the offset correction module as the Smooth L1 loss between the predicted offset $O_{pred}$ and the ground-truth offset $O_{gt}$, formulated as follows:
\begin{equation}
\label{loss}
L_{offset} = \text{SMOOTHL1}\left(O_{pred}, O_{gt}\right).
\end{equation}

%% reviwer 1 q1
The ground truth data is utilized exclusively in the training phase of the offset correction module. Specifically, the module is trained to predict the offset between the ground truth annotations and the detector outputs, allowing it to learn systematic error patterns inherent in the detector. However, at inference time, the ground truth data is not available. Instead, the trained offset correction module directly predicts and rectifies the offsets based solely on the detection results provided by the 3D object detector.
\allowdisplaybreaks
% \vspace{-5pt}
\subsection{Feature Extraction Module}
The feature extraction module uses a combination of information from two important parts: the detected state after the offset correction module and the predicted state from the motion prediction module. This combination helps create feature maps that highlight important characteristics of the object being targeted, using data from the 3D point cloud within the bounding box.
% \begin{figure}[t] %H为当前位置，!htb为忽略美学标准，htbp为浮动图形
% \centering %图片居中
% \includegraphics[width=0.5\textwidth]{feature_extraction.pdf} %插入图片，[]中设置图片大小，{}中是图片文件名
% \caption{The structure of the feature extraction module. The 3D point cloud is served as input into the transformation network (T-Net), where they undergo augmentation before being fed into the PointNet architecture. Following this, the outputs of the penultimate layer of the PointNet are channelled into the multi-layer perception (MLP). This subsequent step yields the desired outputs: the prediction node feature and the detection node feature.} %最终文档中希望显示的图片标题
% \label{feature} %用于文内引用的标签
% \end{figure}

Specifically, the 3D point clouds within the detected and predicted 3D bounding boxes are cropped, feed into a PointNet encoder \cite{qi2017pointnet} to generate corresponding feature maps, denoted as $\bm{F}^{det}_{3d} = \left(\bm{f}^{det}_{3d_{1}},\bm{f}^{det}_{3d_{2}}, \dots, \bm{f}^{det}_{3d_{J}}\right)$ and $\bm{F}^{pred}_{3d} = \left(\bm{f}^{pred}_{3d_{1}},\bm{f}^{pred}_{3d_{2}}, \dots, \bm{f}^{pred}_{3d_{U}}\right)$. Sequentially, the 3D point cloud feature maps are fed into the MLP-based embedding network, to extract embedded features. This process can be described as
\begin{equation}
\label{7}
\mathbf{F}=\mathtt{MLP}\left(\bm{F}^{det}_{3d}\right),
\end{equation}
\begin{equation}
\label{7}
\mathbf{F^*}=\mathtt{MLP}\left(\bm{F}^{pred}_{3d}\right),
\end{equation}
where $\mathbf{F}$ and $\mathbf{F^*}$ denote the output feature maps of the feature extraction module.

\subsection{AMGN Score Calculation Module}
% reviewer2 q2
Utilizing the feature map extracted from the feature extraction module, an AMGN score matrix, denoted as $\bm{B}$, is calculated through the AMGN score calculation module. Like \cite{Probabilistic3DMM}, the purpose of employing such a matrix is to measure the similarity between different detections and predictions. This section details the AMGN Score Calculation Module in three parts: graph construction, AMGN blocks, and the loss function.

\subsubsection{Graph Construction}

Most recent works, which applied GNN for 3D MOT, e.g., PTP \cite{ptp}, Rethinking3DMOT \cite{Rethinking3DMOT}, 3Dmotformer \cite{3dmotformer}, use a fully connected graph with the dimension of $\left(J+U\right) \times \left(J+U\right)$ where $J$ and $U$ denote the number of detected nodes and the number of predicted nodes in the graph, respectively. However, such a graph entails redundant computations due to the simultaneous inclusion of detection and prediction nodes.

To this end, following the approach of GNN3DMOT \cite{Gnn3dmot}, the prior knowledge that matching should only occur across frames is utilized, and a bipartite graph is constructed. The structure of the graph is represented as a matrix of dimensions $J \times U$. The bipartite graph mitigates this computational redundancy, effectively reducing the dimension of the association score matrix. Formally, the bipartite graph is represented as $\bm{G}=\langle \bm{N},\bm{E} \rangle$, with the nodes $\bm{N}$ are split into two distinct sets: one for detected objects $\bm{N}_{det}$ and one for predicted objects $\bm{N}_{pre}$. The set $\bm{E}$ of edges captures the associations between the nodes. 

\begin{figure}[t]
\centering
\includegraphics[width=0.49\textwidth]{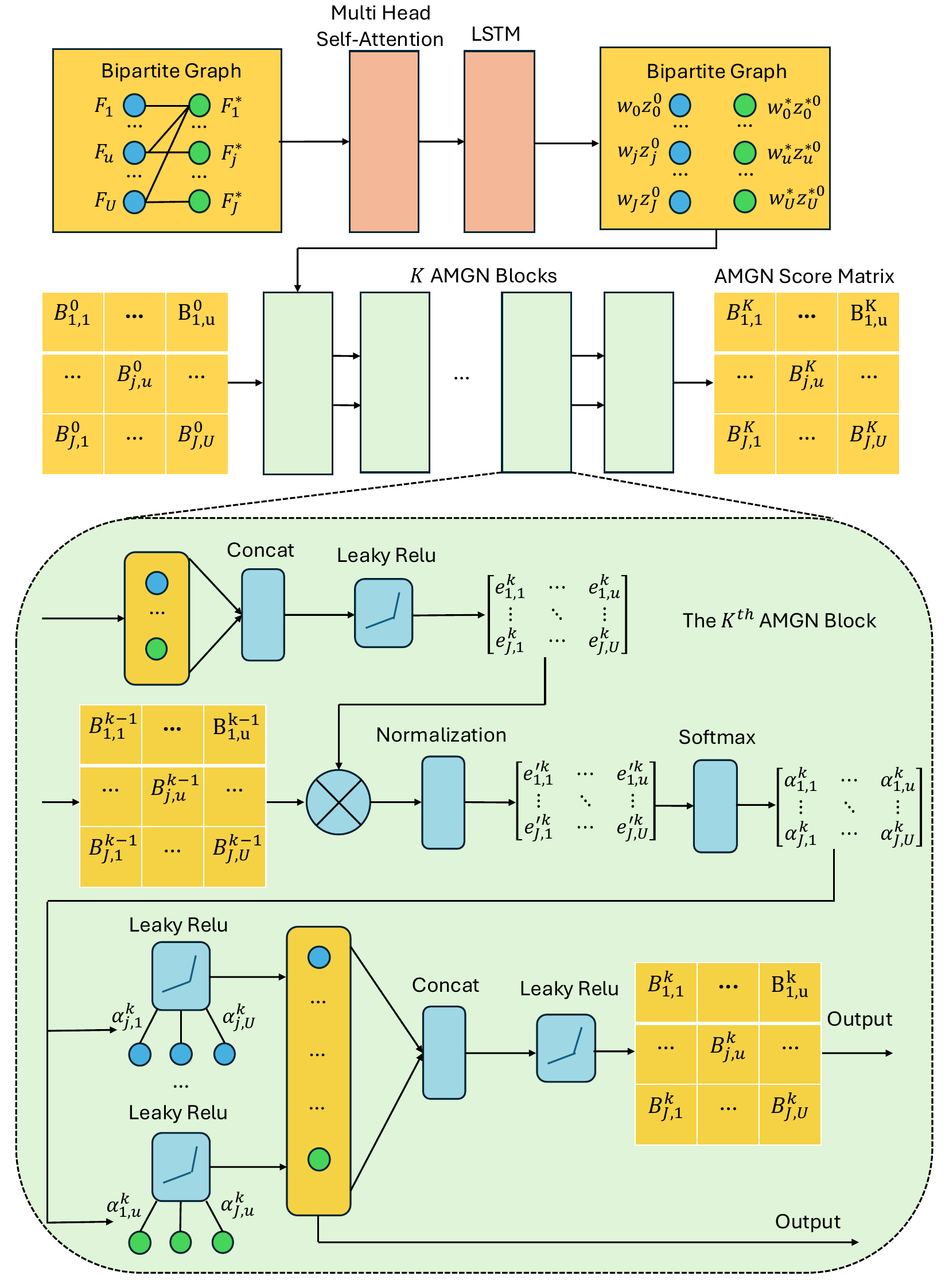}
\caption{AMGN score calculation module. In the initial stage, a bipartite graph is fed into the multi-head self-attention (MHSA) block along with an LSTM block to extract features. This data, together with the initialization of the learnable adjacency matrix, are fed into the AMGN block. The AMGN block performs node message passing and updating using the self-attention coefficients derived from the learnable adjacency matrix. Subsequently, the learnable adjacency matrix is updated with the new feature representation obtained after the message passing and updating process. The new feature representation and updated learnable adjacency matrix are then fed into the next AMGN block. The final output is the AMGN score matrix $\bm{B}$, which is the latest learnable adjacency matrix from the last AMGN block.}
\label{GNN}
\end{figure}
% Sequentially, the initial AMGN score matrix $\bm{B}_{init}$, along with the predicted feature map and detected feature map, are processed as inputs, generating a AMGN score matrix $\bm{B} \in \mathbb{R}^{J \times U}$ as the output, which encodes the importance of each pairwise similarity in the final score.

\subsubsection{AMGN Blocks}

GNN3DMOT \cite{Gnn3dmot} leverages the Graph Attention Network (GAT) \cite{velivckovic2017graph} to enable different contributions from individual nodes during message passing. However, GAT processes nodes independently, and thus it may fail to capture important contextual information embedded in the global graph structure.

To overcome this, the AMGN block is proposed, which introduces a learnable adjacency matrix into the self-attention calculation. This matrix contains the structural information of the graph, allowing the model to continuously refine its understanding of the graph structure, adapt to the evolving feature representations of the nodes, and learn the strength of associations between any pair of nodes. This approach ensures that the attention mechanism incorporates both the node features and the global structural information of the graph, as captured by the learnable adjacency matrix.

The module's process begins by generating robust initial embeddings, which are then iteratively refined by K AMGN blocks. First, a Multi-Head Self-Attention (MHSA) encoder extracts preliminary features from the concatenated detected and predicted feature maps. An LSTM network then processes these features to enhance temporal coherence, generating refined initial embeddings:
\begin{equation}
\label{z0}
    \mathbf{Z}^{0} = \mathtt{LSTM}(\mathtt{MHSA}([\mathbf{F}, \mathbf{F}^*])),
\end{equation}
where $\bm{Z^{0}}$ is the bipartite graph with the embedding representation sets of detected nodes and predicted nodes that contains $[\bm{z}^{0}_{0}, \dots, \bm{z}^{0}_{j}, \dots, \bm{z}^{0}_{J}]$ and $[\bm{z}^{*0}_{0}, \dots, \bm{z}^{*0}_{u}, \dots, \bm{z}^{*0}_{U}]$.

After obtaining the initial embeddings, the K AMGN blocks iteratively refine these embeddings and update the learnable adjacency matrix as shown in Fig. \ref{GNN}. The learnable adjacency matrix is initialized as $\bm{B}^{0}$, where its element $B^{0}_{j,u}$ between the $j$-th node in the set of nodes for detected objects and the $u$-th node in the set of nodes for predicted objects.

For the k-th AMGN block, a raw attention coefficient $e^{k}_{j,u}$ between $j$-th detected node and $u$-th predicted node is first calculated as:
\begin{equation}
\label{cofficient}
e^{k}_{j,u} = \mathtt{LeakyReLU}(a^{T}[w^{*}_{u}z^{*k-1}_{u}||w_{j}z^{k-1}_{j}]),
\end{equation}
where $a$ is the learnable weight vector, $||$ denotes concatenation, and $w^{*}_{u}, w_{j}$ are elements in a learnable weight matrix.

Different from GAT, these attention coefficients are modified by combining them with the learnable adjacency matrix $\bm{B}^{k-1}$ from the previous block as shown:
\begin{equation}
\label{combine}
e^{’k}_{j,u} = e^{k}_{j,u}B^{k-1}_{j,u},
\end{equation}where $B^{k-1}_{j,u}$ is the association score between $j$-th detected node and $u$-th predicted node from the prior step.

The modified coefficients are then normalized using a softmax function as:
\begin{equation}
\label{norm}
\alpha^{k}_{j,u} = \frac{\mathrm{exp}(e^{’k}_{j,u})}{\sum \mathrm{exp}(e^{’k}_{j,u})}.
\end{equation}

With these normalized attention coefficients, the node embedding is updated by aggregating neighbour information. The updated embedding for $j$-th detected node is calculated as:
\begin{equation}
\label{norm2}
z^{’k}_{j} = \mathtt{LeakyReLU}(\sum a_{j,u} w_{u}{z^{*k}_{u}}).
\end{equation}

We do the same for the $u$-th predicted node to get the updated feature representation as $z^{’*k}_{u}$.

Finally, the value in the learnable adjacency matrix $\bm{B}^{k}$ is updated as shown:
\begin{equation}
\label{final}
B^{k}_{j,u} = \mathtt{LeakyReLU}(\mathbf{b}^{T}[z^{’k}_{j}||z^{’*k}_{u}]),
\end{equation}
where $\mathbf{b}^{T}$ represents the learnable weight vector at the $k$-th module. The output of the $k$-th AMGN block serves as the input to the subsequent $(k+1)$-th AMGN block. Ultimately, the AMGN score matrix $\bm{B}$ is defined as $\bm{B}^{K}$, which is the output of the final AMGN block.

\subsubsection{Loss Function}

The corresponding loss function for training the network to generate AMGN score matrix $\bm{B}$, association matrix loss, is expressed mathematically as 
\begin{equation}
\label{12}
L_{a} = \sum_{j}^{J}\sum_{u}^{U} \left[-y_{j,u}\log B_{j,u} -\left(1-y_{j,u}\right)\log\left(1-B_{j,u}\right)\right].
\end{equation}
This formula represents a cross-entropy loss that measures the difference between the predicted association scores ($B_{j,u}$) and the ground truth of the matching pair ($y_{j,u}$). Beware that $y_{j,u}$ is a binary indicator, which is either $0$ or $1$, as it represents the ground truth of the match between the $u$-th predicted node and the $j$-th detected node. When there is a clear match, $y_{j,u}$ is assigned as 1, and 0 otherwise. Due to the absence of ground-truth annotations for each pair, a match is determined based on the continuity of object identity across time frames. In the match, a pair is considered as matched if the closest ground-truth box to the tracking box in the previous time frame and the closest ground-truth box to the detection box in the current time frame have the same index and their IoU with the closest ground-truth box is more than 0.7. The value of $B_{j,u}$ is obtained from the AMGN score matrix $\bm{B}$ and ranges between 0 and 1.

\subsection{Association Module}

In 3D MOT, the data association module plays a crucial role in determining the correspondence between detected and predicted objects, to update the state of the objects. Most of the methods used in 3D MOT rely on the minimization of the geometry association cost matrix. This is done by evaluating the overlap or distance between the 3D bounding boxes of the predicted and detected objects, using the global nearest neighbor principle and typically the Hungarian algorithm  \cite{blackman1999design}. Recently, \cite{PC3T} proposed to incorporate geometry, motion and appearance features from images to construct such association cost matrix. In our proposed approach, we follow a similar idea to define the geometry and motion cost matrix, but using AMGN score matrix $\bm{B}$ to replace the appearance features. The details are discussed in the follow parts.

Same as \cite{PC3T}, the combination of geometry and motion cost matrix is expressed by the mathematical formulation
\begin{equation}
\label{16}
\bm{A} = \bm{Ge} + \bm{Mo},
\end{equation}where the matrix $\bm{Ge}$ denotes a geometry similarity matrix. This $\bm{Ge}$ matrix establishes the correlations between the detected state after the offset correction module, denoted as $\bm{Y}_t^{cor}$, and the predicted state $\bm{X}_{t|t-1}$, via: 
\begin{equation}
\label{17}
\bm{Ge} = \mathtt{IoU}\left(\bm{Y}_t^{cor}, \bm{X}_{t|t-1}\right) + \mathtt{CEN}\left(\bm{Y}_t^{cor}, \bm{X}_{t|t-1}\right).
\end{equation}
The $\mathtt{IoU}$ in Eq. (\ref{17}) is the cost from 3D IoU, which can be computed as
% which has two parts based on the calculation of 3D IoU and the cost from the centroid (in 3D Cartesian coordinate). Here, the 3D IoU is defined as Eq. (\ref{iou}). 
\begin{equation}
\label{iou}
\mathtt{IoU} = \frac{\mathtt{VI}\left(\bm{Y}_t^{cor}, \bm{X}_{t|t-1}\right)}{\mathtt{VOL}\left(\bm{Y}_t^{cor}\right) + \mathtt{VOL}\left(\bm{X}_{t|t-1}\right) - \mathtt{VI}\left(\bm{Y}_t^{cor},\bm{X}_{t|t-1}\right)},
\end{equation}where $\mathtt{VOL}$ is the volume calculated by $w,h,l$ in the state and $\mathtt{VI}$ denotes the volume intersection. The $\mathtt{CEN}$ in Eq. (\ref{17}) is the cost from the centroid (in 3D Cartesian coordinate) \cite{PC3T} which can be computed as
\begin{equation}
\label{cen}
\mathtt{CEN}\left(\bm{Y}_t^{cor}, \bm{X}_{t|t-1}\right) = \mathtt{MSE}\left(\bm{p},\bm{\hat{p}}\right),
\end{equation}where $\bm{p}$ refers to the global coordinates ($x_{p},y_{p},z_{p}$) for detected state (after the offset correction module) $Y_t^{cor}$ and $\bm{\hat{p}}$ is the corresponding value of predicted state $X_{t|t-1}$.

\begin{equation}
\label{18}
{Mo}_{u,j} = w_{ang}*\left(1-\cos\left<\tilde{v},\hat{v}\right>\right)+w_{velo}*\mathtt{MSE}\left(\tilde{v},\hat{v}\right),
\end{equation}where $w_{ang}$ is the weight for angle difference, $w_{velo}$ is the weight for velocity difference, whereas $\tilde{v}$ and $\hat{v}$ denote the velocities of the predicted object states and detected object states respectively.

However, the current approach to calculating the matrix $\bm{A}$ is limited to considering only geometry and motion costs, neglecting the valuable information embedded in the 3D features present within the 3D point cloud. To address this limitation, we propose the final association cost matrix $\bm{C}$, which incorporates the geometry and motion cost matrix $\bm{A}$ and the AMGN score matrix $\bm{B}$, respectively. This amalgamation is achieved through a weighted linear combination of $\bm{A}$ and $\bm{B}$:

% However, the matrix $\bm{A}$ is calculated only based on the geometry cost and motion cost. This constrained approach tends to overlook the indispensable data encapsulated in 3D features embedded within the 3D point cloud. To solve this problem, the final association cost matrix, $\bm{C}$ is established by the amalgamation of the geometry and motion cost matrix $\bm{A}$, and the AMGN score matrix $\bm{B}$. This synthesis is achieved via a weighted linear combination of $\bm{A}$ and $\bm{B}$

\begin{equation}
\label{15}
\bm{C} = \bm{A}-w_{B}\bm{B},
\end{equation}where $w_{B}$ represents the weight assigned to the AMGN score matrix $\bm{B}$. Notably, the subtraction operation between matrix $\bm{A}$ and $\bm{B}$ in Eq. (\ref{15}) is performed due to the score being determined as a negative cost. By incorporating the AMGN score matrix alongside the geometry and motion cost matrix, our proposed method enriches the data association process, allowing for a more comprehensive evaluation of 3D features. With the final association cost matrix $\bm{C}$, as input, the Hungarian algorithm is used to solve the 2D assignment problem, identifying the matched pairs and unmatched objects.

\section{Experiments and Analysis}
\label{experiment}
In this section, we discuss the results of our experiment. We will begin by providing an overview of the dataset used, followed by a description of the implementation details. Next, we will present a thorough ablation study. Lastly, we will compare our results with those of the baseline models.

%-------------------------------------------
\begin{table*}[ht]
%\small
\caption{Comparison of the proposed method and other state-of-the-art LiDAR-only trackers on front view 2D MOT tracking results using KITTI car test dataset.}% \vspace{-15pt}}
\label{comparison_with_sota_lidar_trackers}
\renewcommand\tabcolsep{4pt}
\begin{center}
\begin{threeparttable}
\begin{tabular}{ l|c|cccccccc}
  \hline
  Method & Modality & HOTA(\%)$\uparrow$ & AssA(\%)$\uparrow$ & LocA(\%)$\uparrow$ & MOTA(\%)$\uparrow$ & MOTP(\%)$\uparrow$ & MT(\%)$\uparrow$ & IDS$\downarrow$ & FRAG$\downarrow$ \\
\hline
AB3DMOT (IEEE IROS 2020)* \cite{AB3DMOT} & L & 69.99 & 69.33 & 86.85 & 83.61 & 85.23 & 66.92 & 113 & 206 \\
PC3T (IEEE T-ITS 2022)*$^\#$ \cite{PC3T} & L & 77.80 & \textbf{81.59} & 86.07 & 88.81 & 84.26 & 80.00 & 225 & 201 \\
Batch3DMOT (IEEE RA-L 2022)*
\cite{Batch3DMOT} & L & $N/A$ & $N/A$ & $N/A$ & 88.60 & 86.80 & 76.70 & \textbf{19} & 74 \\
LEGO (\textbf{Ours})* & L & \textbf{78.05} & 79.22 & \textbf{88.08} & \textbf{88.97} & \textbf{86.92} & \textbf{80.92} & 286 & \textbf{71} \\
%LEGO (\textbf{Ours})*^,^\# &  &  &  &  &  &  &  &  \\
\hline
PolarMOT (ECCV 2022)** \cite{PolarMOT} & L & 75.16 & 76.95 & 87.12 & 85.08 & 85.63 & 80.92 & 462 & 599 \\
CenterTube (IEEE T-MM 2023)*** \cite{CenterTube} & L & 71.25 & 69.24 & 86.85 & 86.97 & 85.19 & 78.46 & \textbf{191} & 344 \\
LEGO (\textbf{Ours})** & L & \textbf{79.52} & \textbf{83.34} & \textbf{87.49} & \textbf{88.14} & \textbf{86.06} & \textbf{87.54} & 290 & \textbf{117} \\
\hline
UG3DMOT (Signal Processing 2024)**** \cite{UG3DMOT} & L & 78.60 & 82.28 & 87.84 & 87.98 & 86.56 & 79.08 & \textbf{30} & 360 \\
LEGO (\textbf{Ours})**** & L & \textbf{80.75} & \textbf{83.27} & \textbf{87.92} & \textbf{90.61} & \textbf{86.66} & \textbf{87.85} & 214 & \textbf{109} \\
\hline
%PC3T (IEEE T-ITS 2022)****^,^\# \cite{PC3T} & L & 81.00 & 84.22 & 87.49 & 91.91 & 86.08 & 86.77 & 4.00 & 24 & 107 \\
%LEGO (\textbf{Ours})****^,^\# & L &  &  &  &  &  &  &  &  \\
%\hline
\end{tabular}
\begin{tablenotes}
        \footnotesize
        \item[*] The metrics are reported by using PointRCNN \cite{PointRCNN} as 3D object detector.
        \item[**] The metrics are reported by using PointGNN \cite{PointGNN} as 3D object detector.
        \item[***] The metrics are reported by using CenterPoint \cite{CenterPoint} as 3D object detector, which has much better detection performance than PointGNN in general.
        \item[****] The metrics are reported by using CasA \cite{CasA} as 3D object detector.
        \item[$^\#$] Note: Since the reported metrics of PC3T are based on the trajectory refinement, which makes PC3T an offline smoother rather than online tracker like all others, thus actual performance of online tracking version of PC3T is lower. 
\end{tablenotes}
\end{threeparttable}
\end{center}
%\vspace{-20pt}
\end{table*}

\begin{table*}[ht]
%\small
\caption{\centering{Comparison of the proposed method and other state-of-the-art LiDAR and camera fusion-based trackers on front view 2D MOT tracking results using KITTI car test dataset.}} % \vspace{-15pt}}
\label{comparison_with_sota_lidar_and_camera_fusion_trackers}
\renewcommand\tabcolsep{4pt}
\begin{center}
\begin{threeparttable}
\begin{tabular}{ l|c|cccccccc}
  \hline
  Method & Modality & HOTA(\%)$\uparrow$ & AssA(\%)$\uparrow$ & LocA(\%)$\uparrow$ & MOTA(\%)$\uparrow$ & MOTP(\%)$\uparrow$ & MT(\%)$\uparrow$ & IDS$\downarrow$ & FRAG$\downarrow$ \\
\hline
JMODT (IEEE IROS 2021) \cite{JMODT} & C+L & 70.73 & 	68.76 & 86.95 & 85.35 & 85.37 & 77.39 & 350 & 693 \\
DeepFusionMOT (IEEE RA-L 2022)* \cite{DeepFusionMOT} & C+L & 75.46 & \textbf{80.05} & 86.70 & 84.63 & 85.02 & 68.61 & 84 & 472 \\
StrongFusionMOT (IEEE S-J 2022)* \cite{StrongFusionMOT} & C+L & 75.65 & 79.84 & 86.74 & 85.53 & 85.07 & 66.15 & \textbf{58} & 416 \\
%DFR-FastMOT (Arxiv 2023)* \cite{DFR-FastMOT} &  &  &  &  &  &  \\
Feng et al., (IEEE T-IV 2024)* \cite{Feng} & C+L & 74.81 & $N/A$ & $N/A$ & 84.82 & 85.17 & $N/A$ & $N/A$ & $N/A$ \\
LEGO (\textbf{Ours})* & L & \textbf{78.05} & 79.22 & \textbf{88.08} & \textbf{88.97} & \textbf{86.92} & \textbf{80.92} & 286 & \textbf{71} \\
\hline
EagerMOT (IEEE ICRA 2021)** \cite{EagerMOT} & C+L & 74.39 & 74.16 & 87.17 & 87.82 & 85.69 & 76.15 & 239 & 390 \\
MSA-MOT (Sensors 2022)** \cite{MSA-MOT} & C+L & 78.52 & 82.56 & 87.00 & 88.01 & 85.45 & 86.77 & \textbf{91} & 428 \\
IMSF MOT (IEEE T-ITS 2023) \cite{IMSF-MOT} & C+L & 72.44 & 68.02 & $N/A$ & \textbf{90.32} & 85.47 & 86.46 & 526 & 270 \\
DualTracker (IEEE T-IV 2023)** \cite{DualTracker} & C+L & 74.24 & $N/A$ & $N/A$ & 88.05 & 85.6 & 80.77 & 148 & $N/A$ \\
HIDMOT (IEEE T-VT 2023)** \cite{HIDMOT} & C+L & 75.90 & 77.22 & $N/A$ & $N/A$ & $N/A$ & $N/A$ & $N/A$ & $N/A$ \\
%CAMO-MOT (Arxiv 2022)** \cite{CAMO-MOT} & 79.95 & $N/A$ & $N/A$ & 90.38 & 85.00 & 84.61 & 23 & $N/A$  \\
%DFR-FastMOT (Arxiv 2023)** \cite{DFR-FastMOT} &  &  &  &  &  &  \\
LEGO (\textbf{Ours})** & L & \textbf{79.52} & \textbf{83.34} & \textbf{87.49} & 88.14 & \textbf{86.06} & \textbf{87.54} & 290 & \textbf{117} \\
\hline
MMF-JDT (IEEE RA-L 2025)*** \cite{MMF-JDT} & C+L & 79.52 & \textbf{84.01} & $N/A$ & 88.06 & 86.24 & $N/A$ & \textbf{37} & $N/A$ \\
LEGO (\textbf{Ours})*** & L & \textbf{80.75} & 83.27 & \textbf{87.92} & \textbf{90.61} & \textbf{86.66} & \textbf{87.85} & 214 & \textbf{109} \\
\hline
\end{tabular}
\begin{tablenotes}
        \footnotesize
        \item[*] The metrics are reported by using PointRCNN \cite{PointRCNN} as 3D object detector.
        \item[**] The metrics are reported by using PointGNN \cite{PointGNN} as 3D object detector.
        \item[***] The metrics are reported by using CasA \cite{CasA} as 3D object detector.
\end{tablenotes}
\end{threeparttable}
\end{center}
%\vspace{-20pt}
\end{table*}

\begin{table*}[ht]
%\small
\caption{3D MOT tracking results of the proposed method and other state-of-the-art trackers on KITTI car validation dataset, by following the evaluation protocol in \cite{AB3DMOT}.} % \vspace{-15pt}}
\label{comparison_with_sota_trackers_3D_metric}
\renewcommand\tabcolsep{4pt}
\begin{center}
\begin{threeparttable}
\begin{tabular}{ l|c|ccccccc}
  \hline
  Method & Modality & sAMOTA(\%) & AMOTA(\%)$\uparrow$ & AMOTP(\%)$\uparrow$ & MOTA(\%)$\uparrow$ & MOTP(\%)$\uparrow$ & IDS$\downarrow$ & FRAG$\downarrow$\\
\hline
AdaptiveNoiseCov (IEEE T-IV 2024)$^+$ \cite{AdaptiveNoiseCov} & L & 93.03 & 45.22 & 61.79 & 86.18 & 64.11 & \textbf{0} & 30 \\
AB3DMOT (IEEE IROS 2020)* \cite{AB3DMOT} & L & 93.28 & 45.43 & 77.41 & 86.24 & 78.43 & \textbf{0} & 15   \\
ACK3DMOT (IEEE T-IV 2023)* \cite{ACK3DMOT} & L & $N/A$ & $N/A$ & $N/A$ & 88.73 & \textbf{86.81} & 8 & 68 \\
ConvUKF (IEEE T-IV 2024)* \cite{ConvUKF} & L & 93.32 & 45.46 & 78.09 & $N/A$ & $N/A$ & $N/A$ & 17 \\
FGO-based3DMOT (IEEE S-J 2024)* \cite{Real_Time_MOT} & L & 93.77 & 46.14 & 77.85 & 86.53 & 79.00 & 1 & $N/A$ \\
GNN3DMOT (IEEE CVPR 2020)* \cite{Gnn3dmot} & C+L & 93.68 & 45.27 & 78.10 & 84.70 & 79.03 & \textbf{0} & 10 \\
LEGO (\textbf{Ours})* & L & \textbf{94.90} & \textbf{47.78} & \textbf{86.97} & \textbf{91.36} & 86.70 & 1 & \textbf{4} \\
\hline
PolarMOT (ECCV 2022)** \cite{PolarMOT} & L & 94.32 & $N/A$ & $N/A$ & 93.93 & $N/A$ & 31 & $N/A$ \\
%Intertrack (CRV 2023)** \cite{Intertrack} & L & 95.20 & \textbf{48.80} & 80.30 & $N/A$ & $N/A$ & $N/A$ & $N/A$ \\
CenterTube (IEEE T-MM 2023)*** \cite{CenterTube} & L & 93.89 &  46.24 & 80.23 & $N/A$ & $N/A$ & 78 & $N/A$ \\
EagerMOT (IEEE ICRA 2021)** \cite{EagerMOT} & C+L & 94.94 & \textbf{48.80} & 80.40 & \textbf{96.61} & 80.00 & 2 & $N/A$ \\
HIDMOT (IEEE T-VT 2023)** \cite{HIDMOT} & C+L & $N/A$ & 45.64 & 79.68 & 90.45 & 81.44 & $N/A$ & $N/A$ \\
CAMO-MOT (IEEE T-ITS 2023)** \cite{CAMO-MOT} & C+L & 95.20 & 48.04 & 81.48 & $N/A$ & $N/A$ & $N/A$ & $N/A$ \\
Qiao et al., (IEEE IOT-J 2024)** \cite{AdverseWeather-MOT} & C+L & $N/A$ & 48.35 & 79.90 & 86.33 & 79.42 & 1 & $N/A$ \\
LEGO (\textbf{Ours})** & L & \textbf{95.20} & 48.10 & \textbf{87.05} & 92.00 & \textbf{86.69} & \textbf{1} & \textbf{5} \\
\hline
\end{tabular}
\begin{tablenotes}
        \footnotesize
        \item[$^+$] The metrics are reported by using 3D clustering approach rather than any learning-based model as 3D object detector, according to \cite{AdaptiveNoiseCov}.
        \item[*] The metrics are reported by using PointRCNN \cite{PointRCNN} as 3D object detector.
        \item[**] The metrics are reported by using PointGNN \cite{PointGNN} as 3D object detector.
        \item[***] The metrics are reported by using CenterPoint \cite{CenterPoint} as 3D object detector, which has much better detection performance than PointGNN in general.
\end{tablenotes}
\end{threeparttable}
\end{center}
%\vspace{-20pt}
\end{table*}

\begin{table*}[ht]
%\small
\caption{3D MOT tracking results of the proposed method and other state-of-the-art trackers on Waymo \cite{Waymo} validation dataset.} % \vspace{-15pt}}
\label{comparison_with_sota_trackers_3D_metric_waymo}
\renewcommand\tabcolsep{4pt}
\begin{center}
\begin{threeparttable}
\begin{tabular}{l|c|c|cccc}
  \hline
  Dataset & Method & Modality & MOTA L1(\%)$\uparrow$ & MOTA L2(\%)$\uparrow$ & MOTP L1(\%)$\uparrow$ & MOTP L2(\%)$\uparrow$ \\
\hline
& SimTrack (ICCV 2021)$^+$ \cite{SimTrack} & L & 53.1 & 49.6 & 17.4 & 17.4 \\
& 3DMODT (IEEE ICRA 2023)$^+$ \cite{3DMODT} & L & 55.9 & 51.2 & 18.9 & 18.9    \\
& SimpleTrack (ECCV Workshop 2022)* \cite{SimpleTrack} & L & $N/A$ & 56.9 & $N/A$ & $N/A$  \\
\multirow{1}{*}{Waymo} & LEGO (\textbf{Ours})* & L & \textbf{58.30}  & \textbf{58.30}  & \textbf{19.44}  & \textbf{19.44}  \\
\cline{2-7}
& PC3T (IEEE T-ITS 2022)** \cite{PC3T} & L & 61.3  & 61.3
 & $N/A$ & $N/A$  \\
& Fast-Poly (IEEE RA-L 2024)** \cite{Fast-Poly} & L & 62.3  & 62.3 & $N/A$ & $N/A$  \\
& LEGO (\textbf{Ours})** & L & \textbf{63.1}  & \textbf{63.1}  & \textbf{23.6}  &  \textbf{23.6} \\
\hline

\end{tabular}
\begin{tablenotes}
        \footnotesize
        \item[$^+$] The metrics are reported without using pre-trained 3D object detector, but a joint detection and tracking network delicately for this task.
        \item[*] The metrics are reported by using CenterPoint \cite{CenterPoint} as 3D object detector.
        \item[**] The metrics are reported by using CasA \cite{CasA} as 3D object detector.
\end{tablenotes}
\end{threeparttable}
\end{center}
%\vspace{-20pt}
\end{table*}

\subsection{Dataset and Evaluation Metrics}
 % Waymo \cite{Sun2020ScalabilityIP}, nuScenes \cite{Nuscenes}, and Argoverse \cite{Chang2019Argoverse3T}
In this work, we use the KITTI dataset \cite{Geiger2013VisionMR} and Waymo dataset \cite{Waymodata} as many LiDAR-based MOT methods have been evaluated within these datasets. The efficacy of our proposed LEGO modular tracker is assessed through the Higher-Order Tracking Accuracy (HOTA) \cite{hota}, defined as
\begin{equation}
    HOTA = \int_{0}^{1} HOTA_{\alpha }d\alpha \approx \frac{1}{19} \sum_{\alpha}HOTA_{\alpha },
    \label{hotaa}
\end{equation}

\begin{equation}
    HOTA_{\alpha } = \sqrt{\frac{\sum_{c}\mathcal{A}(c) }{FN+FP+TP} },
    \label{hotaaa}
\end{equation}where $FN$, $FP$, and $TP$ represent the number of false negatives, false positives, and true positives, respectively, and $\mathcal{A}(c)$ is the data association score. In Eq. (\ref{hotaa}), $\alpha \in \left ( 0.05, 0.1, \dots, 0.9, 0.95 \right )$ is a particular localization threshold used to determine false negatives and positives. In addition to HOTA, various other evaluation metrics are employed including Association Accuracy (AssA), Localization Accuracy (LocA), Multiple Object Tracking Accuracy (MOTA), Multiple Objects Tracking Precision (MOTP), Mostly Tracked Trajectories (MT, indicating the proportion of ground-truth trajectories that are at least 80$\%$ covered by the tracking output), Mostly Lost Trajectories (ML, denoting the proportion of ground-truth trajectories that are at most 20$\%$ covered by the tracking output), the quantity of Identity Switches (IDS), and the number of instances a trajectory is Fragmented (FRAG).
The MOTA is defined as
\begin{equation}
    MOTA = 1-\frac{\sum_{t} FN_{t}+FP_{t}+IDS_{t}}{\sum _t GT_{t}} ,
    \label{mota}
\end{equation}
where $FN_{t}$, $FP_{t}$, $IDS_t$, and $GT_{t}$ represent the number of false negatives, false positives, ID switch, and ground truth at time $t$ respectively. The MOTP is defined as
\begin{equation}
    MOTP = \frac{\sum_t dis_{t}}{\sum _t c_{t}},
    \label{motp}
\end{equation}
where $dis_{t}$ represents the distance between detection and its corresponding ground truth, and $c_{t}$ is the number of matched pairs at time $t$. Considering that the evaluation metrics on the KITTI benchmark are predominantly oriented towards a 2D perspective, additional comprehensive metrics pertinent to a 3D viewpoint are also employed, such as the Average Multiple Object Tracking Accuracy (AMOTA) and the corresponding precision metric, the Average Multiple Object Tracking Precision (AMOTP). AMOTA is defined as
\begin{equation}
    AMOTA = \frac{1}{M} \sum_{r}\left(1-\frac{\sum_{t} FN_{t}^{r}+FP_{t}^{r}+IDS_{t}^{r}}{\sum _t GT_{t}}\right),
    \label{amota}
\end{equation}
where $FN_{t}^{r}$, $FP_{t}^{r}$, $IDS_{t}^{r}$ represent the number of false negatives, false positives, ID switch at a specific recall value $r$ at time $t$, $M$ is the number of recall values respectively and AMOTP is defined as
\begin{equation}
    AMOTP = \frac{1}{M} \sum_{r}\left(1-MOTP_{r}\right),
    \label{amotp}
\end{equation}
where $MOTP_{r}$ is the value of MOTP at a specific recall ${r}$.
%----------------------------------------------

\begin{figure*}[htbp]
\centering
    \subfigure[Tracking ground truth for scene 0008]{
    \includegraphics[width=0.33\textwidth]{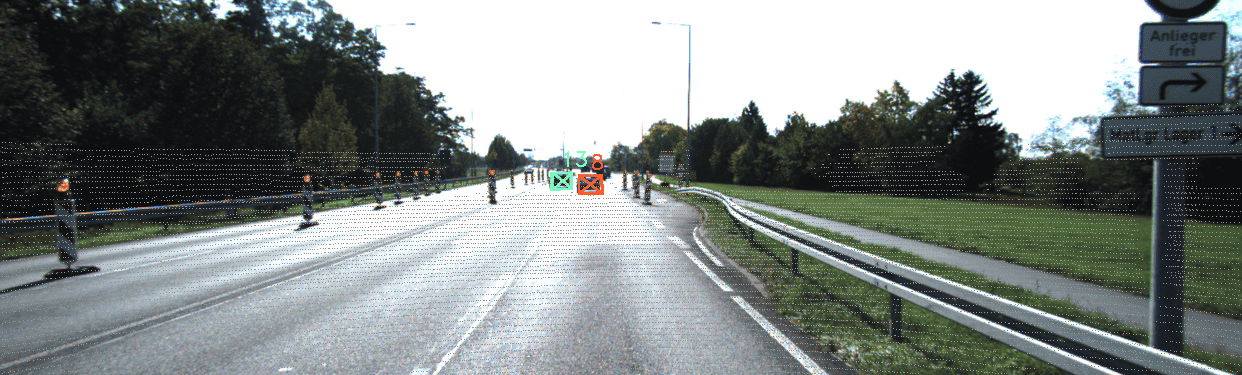}
    \includegraphics[width=0.33\textwidth]{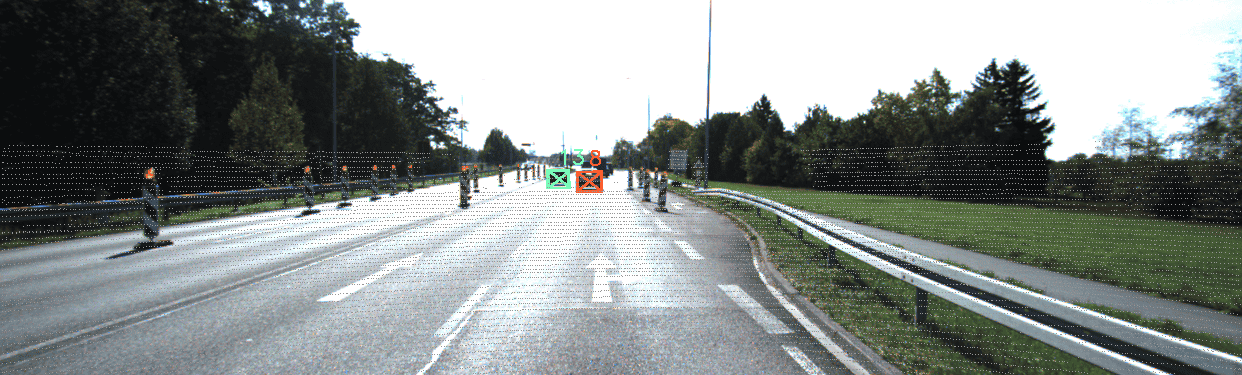}
    \includegraphics[width=0.33\textwidth]{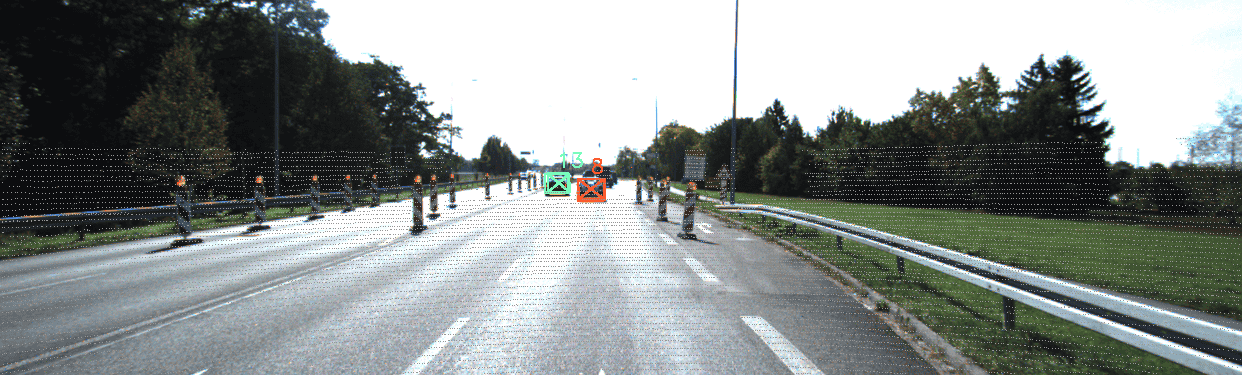}
    }
    \subfigure[PC3T tracking results for scene 0008]{
    \includegraphics[width=0.33\textwidth]{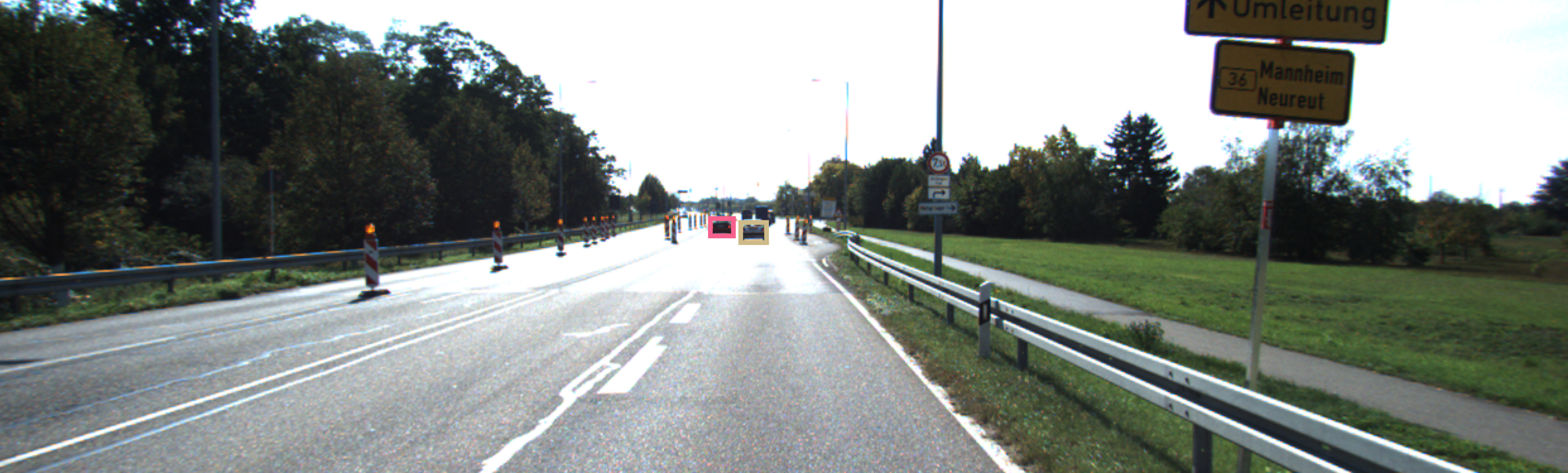}
    \includegraphics[width=0.33\textwidth]{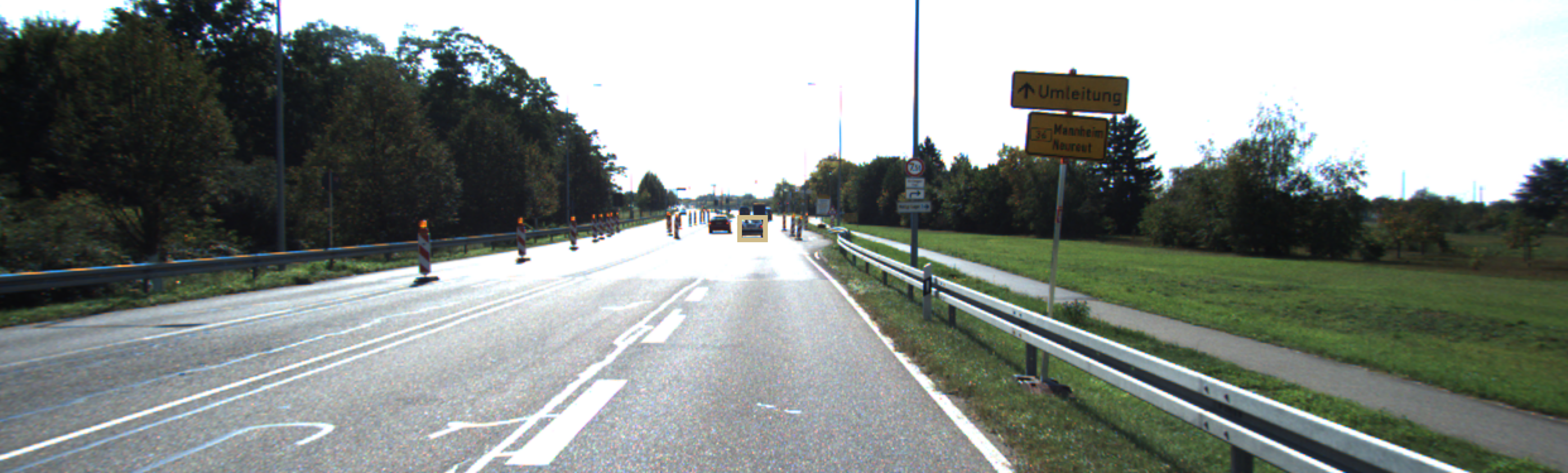}
    \includegraphics[width=0.33\textwidth]{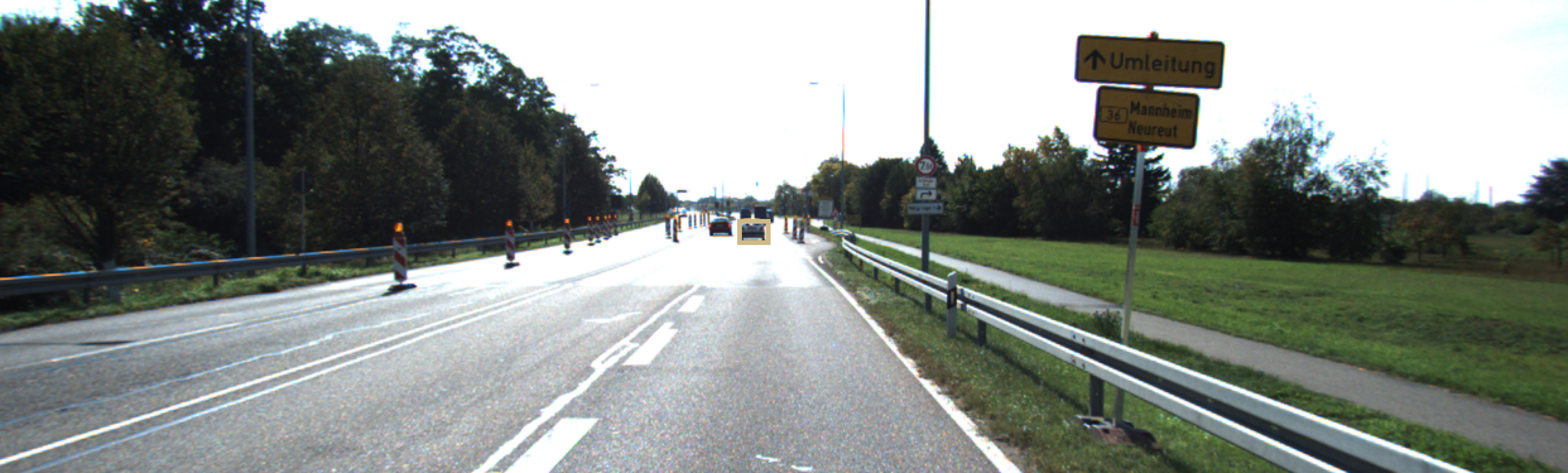}
    }
    \subfigure[LEGO tracking results for scene 0008]{
    \includegraphics[width=0.33\textwidth]{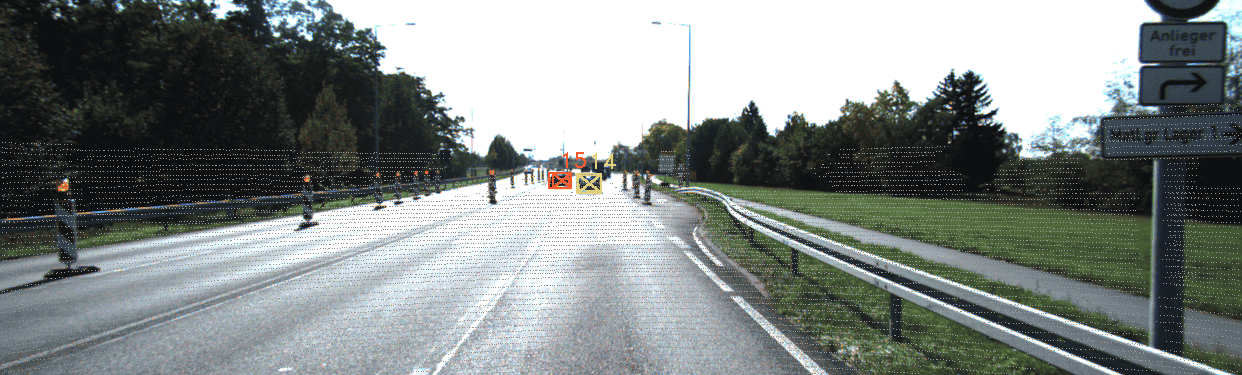}
    \includegraphics[width=0.33\textwidth]{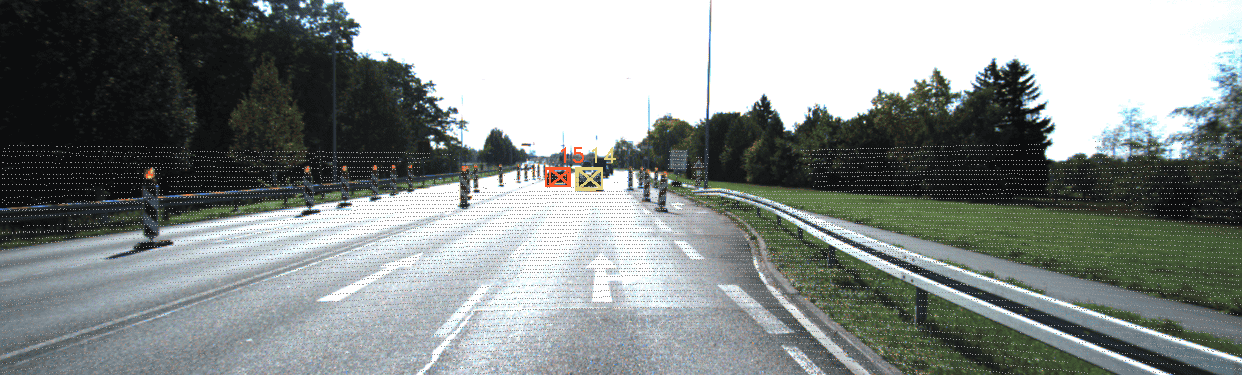}
    \includegraphics[width=0.33\textwidth]{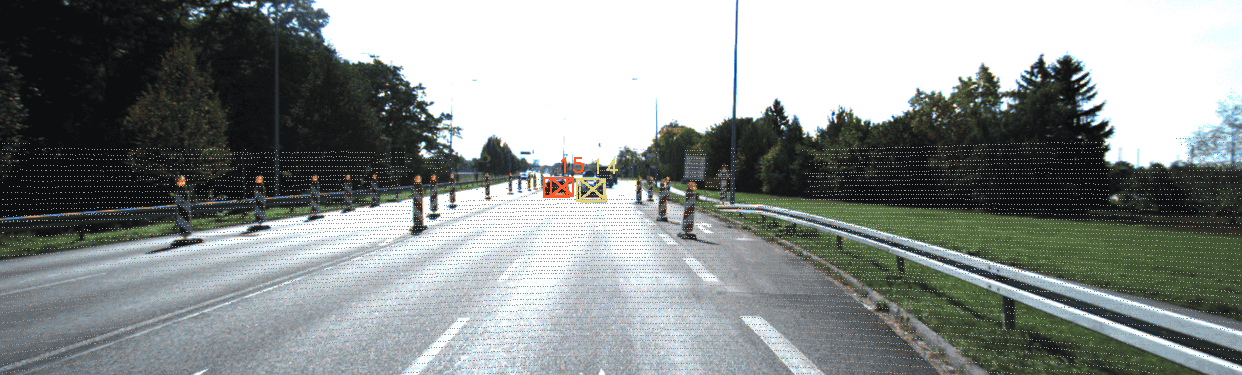}
    }
    \caption{
    Visualization of 3D MOT results on the KITTI benchmark (scene 0008). The first row shows the ground truth tracking annotations across three consecutive frames. Each detected object in the current frame is represented by a bounding box along with its unique tracking ID. Objects are consistently colored according to their tracking IDS across frames. The second and third rows illustrate the tracking results obtained by PC3T and our proposed LEGO model, respectively, on the same frames. Please zoom in to see the details.
    }
    \label{vis}
\end{figure*}

\subsection{Implementation Details}

During the training, Adam with a momentum set as 0.09 and an initial learning rate of 0.1 is applied. The learning rate decay is 0.001, and the training is executed across 100 epochs. During the inference phase, the LiDAR scanning frequency is fixed at 10Hz, and the threshold for the existence probability provided by the object detector is set to 0. This means that all the objects provided by the object detector are kept as input to the tracker. The track management threshold $N_{t}$ and $M_{t}$ are set to 14 and 4, respectively. Implementation details of the key modules are as follows:

\textbf{Offset correction module:} The kernel size, stride, and padding of the CNN layer are set to (3, 1, 1), respectively. The sizes of the two-layer MLP are set to (64, 6). 

\textbf{Feature extraction module:} This module commences by setting the convolution channels to [64, 128, 1024] and configuring the kernel size and stride as 3 and 1, respectively, within the T-Net. Subsequently, a three-layer MLP is structured with sizes of (1024, 256, 128).

\textbf{AMGN score calculation module:} The number of AMGN blocks, $K$, is set to 3. 

\textbf{Time Complexity:} Our method achieves a frame rate of 27.83 FPS on a single NVIDIA RTX 3080 GPU with an Intel(R) Xeon(R) Platinum CPU, measured across the entire KITTI validation dataset. Although direct FPS comparisons with existing methods are challenging due to differences in hardware configurations, our tracking pipeline requires no post-processing operations, thus functioning fully online and in real-time. Compared to previously reported methods such as FANTrack [19] (25.0 FPS), 3DT [72] (33.3 FPS), mmMOT [20] (4.8 FPS), and GNN3DMOT [52] (5.2 FPS). Although the hardware configurations used by these methods were not disclosed, limiting direct comparisons, our method achieves 27.83 FPS, which meets the requirements for online tracking speed.

% The offset extraction module takes in the coordinates, weight, height, and length of detected objects as inputs to predict the offset to the corresponding labelled ground-truths. The cross-entropy loss is used to calculate the difference between the predicted offset and the ground truth during training phase.

% The feature extraction module employs the point cloud within the object's 3D bounding box and corresponding class information as input, generates a feature map from its last two layers. The module uses the cross-entropy loss function for training and has a convolutional size of [64, 128, 1024], with a dropout rate of 0.3 to improve generalization.

% The AMGN score calculation module takes the feature map generated by the feature extraction module as input and estimates an association matrix to represent data associations relationship across time $t$ and $t-1$. The AMGN module has a dropout rate of 0.5 to improve model robustness.

% The Adam algorithm is used for optimization, with a momentum parameter of 0.09 and a learning rate decay of 0.0005. The learning rate is set at 0.001, and the training process is executed for 100 epochs.

\subsection{Performance Comparison with Other State-of-the-Arts}
\label{sotacomp}

In this subsection, the framework is evaluated and discussed concerning various MOT metrics.

\subsubsection{Quality of Detection Input}
The performance of a tracker is inherently tied to the effectiveness of the integrated detector. Three distinct object detectors were assessed in the conducted experiments: the CasA 3D detector, the PointGNN 3D detector, and the PointRCNN 3D detector. These 3D detectors were chosen based on their performance in the KITTI 3D Object detection challenge and arranged in descending order of effectiveness. 
%As depicted in Table \ref{comparison_with_sota_lidar_trackers}, the tracking performances align with the quality of the detection input for all the assessed trackers. 

\subsubsection{Performance Comparison with State-of-the-Art Trackers Using LiDAR Only on KITTI dataset with 2D MOT Metrics}
\label{comparison_lidar_tracker}

The typical tracking result of our LEGO has been visualized in Fig. \ref{vis}. The first row and the second show the ground truth and tracking result reported by our LEGO in scene 0006, the track state for every object in the current frame is marked by a bounding box and its track ID. The third row and the fourth-row show results in scene 0008. As delineated in Table \ref{comparison_with_sota_lidar_trackers}, a series of comparative analyses were conducted on various tracking methodologies within the context of the KITTI tracking benchmark. First, we compare our method with PC3T \cite{PC3T}, utilizing PointRCNN as the object detector. The results reveal that our method achieved an enhancement of 0.25 in HOTA and 0.37 in MOTA on the testing dataset. Further analysis is extended to other methodologies utilizing disparate detectors, as outlined in Table \ref{comparison_with_sota_lidar_trackers}. For instance, in a comparison with CenterTube \cite{CenterTube}, which employs the same detector PointGNN, our method demonstrates an improvement of 8.32 in HOTA and 1.17 in MOTA. Likewise, when compared with UG3DMOT \cite{UG3DMOT} that utilizes the CasA detector, our method achieved an enhancement of 2.15 in HOTA and 2.63 in MOTA.
% Our proposed tracker, LEGO, is contrasted with various detector model-based LiDAR-only trackers. The evaluation results using the KITTI test dataset are presented in Table \ref{comparison_with_sota_lidar_trackers}. Among the trackers that utilised the CasA detector, the LEGO tracker achieved the highest HOTA score of 80.75 in an online scenario. With regards to AssA and LocA metrics, the LEGO tracker scored 83.27 and 87.92, respectively, outperforming UG3DMOT \cite{UG3DMOT}. Furthermore, LEGO obtained higher scores in both MOTA (90.61) and MOTP (86.66) than the UG3DMOT. In addition, our proposed method also excelled in terms of MT, ML, and FRAG.

% As for the PointRCNN detector and PointGNN detector-based trackers, LEGO surpassed the performance of all other trackers using the same detector. Utilizing the PointRCNN, LEGO attained a score of 78.05, an improvement of 0.25 over PC3T. In relation to LocA, MOTA, MOTP, and MT metrics, LEGO outperformed PC3T, with MOTP and MT exceeding by 2.66 and 0.92 respectively. When utilizing the PointGNN, LEGO continued to exhibit superior performance in HOTA, AssA, LocA, MOTA, MOTP, ML, and FRAG, with the only exception being IDS.

\subsubsection{Performance Comparison with State-of-the-Art Trackers Using LiDAR and Camera on KITTI dataset with 2D MOT Metrics} Many tracking algorithms leverage the fusion of 2D camera images and 3D point clouds to optimize performance within the KITTI tracking benchmark. Our proposed method, LEGO, was rigorously evaluated against a variety of LiDAR and camera fusion-based trackers. The comparative analysis was performed in two key segments in Table \ref{comparison_with_sota_lidar_and_camera_fusion_trackers}, based on the detectors employed. Firstly, LEGO was compared with trackers utilizing PointRCNN as their detector. In this context, a notable improvement was observed against StrongFusionMOT \cite{StrongFusionMOT}, with LEGO registering 2.4 increments in HOTA and 3.44 increments in MOTA. The second segment of the comparative analysis involved trackers that employ PointGNN as their detector. LEGO's improvements are again manifest, with a 3.62 enhancement in HOTA over HIDMOT \cite{HIDMOT} and a 5.34 advancement in HOTA and 0.09 advancement in MOTA over DualTracker \cite{DualTracker}. A specific comparison with IMSF-MOT \cite{IMSF-MOT} highlighted LEGO's superior performance.
% In this section, our proposed LEGO tracker is compared with state-of-the-art LiDAR and camera fusion-based trackers in a 2D view. The results are presented in Table \ref{comparison_with_sota_lidar_and_camera_fusion_trackers}. Among the trackers that utilize PointRCNN as the 3D object detector, our LEGO tracker outperformed StrongFusionMOT by 2.35 in terms of HOTA. Although our LEGO tracker scored 0.83 less than DeepFusionMOT and 0.62  less than StrongFusionMOT in AssA, it demonstrated excellent performance in LocA, MOTA, MOTP, and MT. Among the trackers that employ PointGNN as the 3D object detector, our LEGO tracker consistently exhibited superior performance compared to EagerMOT, DualTracker, HIDMOT, and MSA-MOT in terms of HOTA, AssA, LocA, MOTA, MOTP, ML, and FRAG. In comparison with the recently proposed IMSF-MOT, our approach has demonstrated superior performance despite its notable performance with a MOTA of 90.32, higher than our LEGO, and an IDS of 91, which is lower than ours. In particular, our method has shown improved results in several metrics, including HOTA, MT, ML, among others.

\subsubsection{Comparison of Proposed Method and Other State-of-the-Art Trackers on KITTI dataset with 3D MOT Metrics}
Our proposed LEGO tracker has also been subjected to extensive comparative evaluation against various trackers for tracking in 3D space, by employing 3D MOT metrics, such as sAMOTA, AMOTA, and AMOTP. The details of these comparisons are tabulated in Table \ref{comparison_with_sota_trackers_3D_metric}. By employing PointRCNN as 3D detector, LEGO demonstrated the best overall performance across several fronts, comparing to all other recent proposed trackers using LiDAR only. Additionally, LEGO was evaluated against CenterTube \cite{CenterTube} which utilizes CenterPoint as its 3D detector, generally considered to offer superior detection performance over PointGNN. Despite using PointGNN as the 3D detector, LEGO still managed to secure improvements of 1.31 in sAMOTA, 1.86 in AMOTA, and 6.82 in AMOTP. The comparisons were further extended to trackers with fusion of LiDAR and camera. Utilizing PointGNN as 3D detector, when juxtaposed with CAMO-MOT \cite{CAMO-MOT}, LEGO matched the sAMOTA, but exhibited improvement in AMOTA and AMOTP, by 0.06 and 5.57, respectively. Besides, the LEGO system also demonstrates significant enhancements when compared with the recently proposed approach \cite{AdverseWeather-MOT}, achieving a 7.15 improvement in AMOTP and a 5.67 increase in MOTA.

\subsubsection{Comparison of Proposed Method and Other State-of-the-Art Trackers on Waymo Dataset with 3D MOT Metrics}
To demonstrate the generalization capability of our proposed LEGO tracker, the evaluation of LEGO tracker against other state-of-the-art methods on the Waymo dataset is also shown in Table \ref{comparison_with_sota_trackers_3D_metric_waymo}. Using CenterPoint as the detector, LEGO demonstrates superior performance, achieving 58.30 MOTA for both L1 and L2 metrics. This represents a notable improvement of 1.4 over SimpleTrack on L2 MOTA. Furthermore, LEGO significantly outperforms previous methods, surpassing SimTrack and 3DMODT by 5.2 and 2.4 respectively on MOTA L1. LEGO also exhibits better performance with MOTP scores of 19.44 for both L1 and L2 metrics, demonstrating improvements of 2.04 and 0.54 compared to SimTrack and 3DMODT, respectively. When integrated with the CasA detector, LEGO's performance improves with 63.1 MOTA and 23.6 MOTP.
% when utilizing the PointRCNN detector, LEGO achieved an sAMOTA score of 94.9 and an AMOTA score of 47.78. Additionally, LEGO demonstrated a substantial AMTP score that is 8.87 points superior to GNN3DMOT. For the FRAG metric which is lower well, LEGO gets only 4. With the PointGNN detector, LEGO also displayed commendable performance. It achieved the highest sAMOTA score at 95.20 and an AMOTA score of 48.10, which, while slightly lower than EagerMOT, remains competitive. LEGO also recorded the highest AMOTP score of 87.05 and an equally low FRAG of 5. Such results highlight the immense potential for further enhancement of performance by expanding the LEGO tracking framework to incorporate settings that fuse LiDAR and camera data.
% reviewer 2 q4
\subsubsection{Additional Analysis}
Although our proposed LEGO tracker achieves superior HOTA and MOTA scores compared to other state-of-the-art methods (as shown in Table I), we observe an increased number of identity switches (IDS). The AMGN module primarily focuses on relational features derived from the current and immediately preceding frames. Without explicit long-term memory or robust temporal smoothing mechanisms, the tracker may face difficulties in maintaining consistent object identities over extended periods, especially within highly dynamic scenes, further increasing IDS. Moreover, the current absence of appearance-based features limits the stability of associations, exacerbating IDS issues. In future work, these limitations are planned to be addressed by incorporating a lightweight module or knowledge distillation integrated with appearance features, aiming to enhance the stability and robustness of associations and thereby reduce IDS.

\subsection{Ablation Study for LEGO Modular Tracker}
\label{ablation}
This section outlines the ablation study for various key modules in LEGO, all the experiments are implemented on KITTI car validation dataset. 
\subsubsection{Effectiveness of Modules}
Table \ref{ablation} demonstrates the impact of the offset correction and AMGN score
calculation modules on model performance. When both modules are enabled, the model achieves the best results across all metrics, with the HOTA of 85.808$\%$, highlighting their complementary benefits. Enabling only one module improves performance compared to the baseline as well. The AMGN score calculation module alone provides slightly better gains (HOTA: 84.80$\%$) than the offset module alone (HOTA: 83.079$\%$). Disabling both modules results in the lowest performance (HOTA: 82.60$\%$), emphasizing the necessity of these components.
\subsubsection{Effectiveness of Weight between Matrix $\bm{A}$ and Matrix $\bm{B}$}
Throughout the experimental phase, refinements were introduced to the cost weight $w_{B}$ to ascertain the optimal equilibrium between the geometry and motion cost matrix $\bm{A}$, and the AMGN score matrix $\bm{B}$. A comprehensive assessment was conducted across four distinct levels for $w_{B}$, specifically 0, 1, 2, 3, and 4, as described in 
Table \ref{weightmatrix}. % reviewer3 q1 
When $w_{B}$ is set to 0, the AMGN module is disabled, meaning the graph structure and associated learned relationships are not utilized. Consequently, this configuration yields the lowest tracking performance. It was found that when $w_{2}$ was set to 2, the best outcomes were achieved. %This highlights the importance of incorporating detections in maintaining trajectory continuity.
These results underscore the pivotal role played by the cost weight in determining the relative contributions of matrix $\bm{A}$ and matrix $\bm{B}$ to the association process.

\subsubsection{Effectiveness of Parameter $K$ in AMGN Module}
To investigate the effectiveness of parameter $K$ within the proposed AMGN module, an ablation study is conducted on the KITTI car validation dataset. Table~\ref{threshold} summarizes the performance in terms of HOTA, AssA, and LocA across varying values of $K$. It is observed that increasing $K$ from 0 to 3 consistently improves tracking performance, with the best results achieved at K=3, yielding HOTA of 85.808$\%$, AssA of 88.63 $\%$, and LocA of 92.65$\%$. However, further increasing $K$ to 4 slightly decreases performance, indicating diminishing returns and potential overfitting. Based on these findings, the parameter K is set to 3 as the optimal parameter setting for the AMGN module in our subsequent experiments.

\begin{table}[ht]
\caption{Ablation study of proposed modules on the KITTI car validation dataset.}
\label{ablation}
\begin{center}
\resizebox{0.4\textwidth}{!}{
\begin{tabular}{c|c|c|c|c}
\hline
Offset &  AMGN &  HOTA(\%)$\uparrow$ &  AssA(\%)$\uparrow$ & LocA(\%)$\uparrow$ \\ \hline
\ding{55} &\ding{55} & 82.600 & 85.01 & 88.00 \\
\hline

% check this row, see whether the metric here are correct or not. Delete the previous row if this is the suitable metric.
\checkmark & \ding{55} & 83.079 & 85.954 & 90.649  \\
\hline
\ding{55} & \checkmark & 84.800 & 87.20 & 91.00  \\
\hline
\checkmark & \checkmark & \textbf{85.808} & \textbf{88.63} & \textbf{92.65}  \\
\hline
\end{tabular}
}
\end{center}
\end{table}

\begin{table}[ht]
%\small
\caption{Ablation study of the weight $w_{B}$ for matrix $\bm{B}$ with proposed method on KITTI car validation dataset.} % \vspace{-15pt}}
\label{weightmatrix}
\renewcommand\tabcolsep{4pt}
\begin{center}
\begin{threeparttable}
\begin{tabular}{ l|cccc}
  \hline
  $w_{B}$ & HOTA(\%)$\uparrow$ & AssA(\%)$\uparrow$ & LocA(\%)$\uparrow$ & MOTA(\%)$\uparrow$
\\
\hline
0 & 83.079 & 85.954 & 90.649 & 89.271  \\
1 & 83.833 & 88.162 & 92.504 & 85.452  \\
2 & \textbf{85.808} & \textbf{88.63} & \textbf{92.65} & \textbf{90.321}\\
3 & 84.554 & 88.609 & 92.472 & 86.657 \\
4 & 82.242 & 84.19 & 92.492 & 86.418  \\
\hline
\end{tabular}
\end{threeparttable}
\end{center}
%\vspace{-20pt}
\end{table}

\iffalse
\begin{table}[ht]
\caption{Ablation study on the threshold in track management with proposed method on KITTI car validation dataset.} % \vspace{-15pt}}
\label{threshold}
%\setlength{\tabcolsep}{6pt}
%\renewcommand{\arraystretch}{0.7}
\renewcommand\tabcolsep{4pt}
\begin{center}
\begin{threeparttable}
\begin{tabular}{ l|ccc}
  \hline
  $N^{t}$ & HOTA(\%)$\uparrow$ & AssA(\%)$\uparrow$ & LocA(\%)$\uparrow$
\\
\hline
10 & 85.305 & 86.704 & 92.371 \\
12 & 85.522 & 87.045 & 92.35 \\
14 & \textbf{85.808} & \textbf{88.63} & \textbf{92.65} \\
15 & 84.411 & 88.452 & 92.469 \\
24 & 84.189 & 88.198 & 92.457 \\
34 & 83.692 & 87.466 & 92.337 \\
\hline
\end{tabular}
\end{threeparttable}
\end{center}
%\vspace{-20pt}
\end{table}
\fi

\begin{table}[ht]
%\small
\caption{Ablation study on the parameter K in AMGN Module with proposed method on KITTI car validation dataset.} % \vspace{-15pt}}
\label{threshold}
\renewcommand\tabcolsep{4pt}
\begin{center}
\begin{threeparttable}

\begin{tabular}{ l|ccc}
  \hline
  $K$ & HOTA(\%) & AssA(\%)$\uparrow$ & LocA(\%)$\uparrow$
\\
\hline
0 & 82.600 & 85.01 & 88.00 \\
1 & 83.357 & 86.592 & 90.677 \\
2 & 84.263 & 87.014 & 91.588 \\
3 & \textbf{85.808} & \textbf{88.63} & \textbf{92.65} \\
4 & 85.189 & 88.038 & 92.257 \\
\hline
\end{tabular}

\end{threeparttable}
\end{center}
%\vspace{-20pt}
\end{table}

 % Required for checkmark and cross symbols

%\subsubsection{Effectiveness of Threshold in Track Management} In the experiments, the $N^{t}$ parameter was adjusted as 14 to manage tracks. Table \ref{threshold} provides the results obtained by varying the threshold. These findings indicate that an optimal threshold pre-processing operation, combined with the appropriate cost weight used in the proposed LEGO tracker, can improve the HOTA score.

%===============================================
\section{Conclusion}
\label{conclusion}
%The proposed tracker leverages 3D point cloud information, topological structures, and motion prediction features to enhance tracking accuracy and robustness.
In this paper, we propose an online LiDAR-based tracker, LEGO, that introduces the offset correction module and AMGN score calculation module, and effectively integrates them into the existing tracking framework to tackle the inherent challenges of MOT. Our offset correction module demonstrates the capability to rectify certain errors in the detection results, contributing to improved tracking performance. Furthermore, the AMGN blocks which use the learnable adjacency matrix to estimate the association relationships between detected objects and predicted objects in an efficient way. However, it is important to acknowledge that the use of graph structures in our approach may lead to longer training times due to increased computational demands. As a potential solution for future research, techniques, such as voxel downsampling could be implemented to reduce the parameter count in the GNN, thereby enhancing computational efficiency and reducing time consumption.
    
%% ----------------------------------------------------------------------------------------------------

% use section* for acknowledgement
%\section*{Acknowledgment}
%The authors would like to thank...

%\small
\footnotesize
%\bibliographystyle{IEEEtran}
%\bibliographystyle{ieeeconf}
%\bibliography{reference}
% \begin{thebibliography}{60}
%\bibliographystyle{plain} 
%\bibliography{lego}

\begin{thebibliography}{60}

\bibitem{Pedestrian_MOT} N. McLaughlin, J. M. del Rincon and P. Miller, ``Video Person Re-Identification for Wide Area Tracking Based on Recurrent Neural Networks,'' \emph{IEEE Transactions on Circuits and Systems for Video Technology}, vol. 29, no. 9, pp. 2613-2626, 2019.

\bibitem{Pedestrian_MOT_2} H. Nodehi and A. Shahbahrami, ``Multi-Metric Re-Identification for Online Multi-Person Tracking,'' \emph{IEEE Transactions on Circuits and Systems for Video Technology}, vol. 32, no. 1, pp. 147-159, 2022.

\bibitem{Drone_ITS_MOT}
H. Jin, X. Nie, Y. Yan, X. Chen, Z. Zhu and D. Qi, “AHOR: Online Multi-Object Tracking With Authenticity Hierarchizing and Occlusion Recovery,” IEEE Transactions on Circuits and Systems for Video Tech- nology, vol. 34, no. 9, pp. 8253-8265, 2024.

\bibitem{Drone_ITS_MOT_2} H. Wu, H. Sun, K. Ji and G. Kuang, ``Temporal-Spatial Feature Interaction Network for Multi-Drone Multi-Object Tracking,'' \emph{IEEE Transactions on Circuits and Systems for Video Technology}, vol. 35, no. 2, pp. 1165-1179, 2024.

\bibitem{3dsot} J. Xiao, Y. Ma, W. Yang and T. Zhang, “Learning Adaptive Conceptual Prototypes for 3D Single Object Tracking,” IEEE Transactions on Circuits and Systems for Video Technology, 2025.



\bibitem{MOT_ITS} L. Liu et al., ``Yolo-3DMM for Simultaneous Multiple Object Detection and Tracking in Traffic Scenarios,'' \emph{IEEE Transactions on Intelligent Transportation Systems}, vol. 25, no. 8, pp. 9467-9481, Aug. 2024.

%\bibitem{Drone_ITS_MOT} I. Bisio, C. Garibotto, H. Haleem, F. Lavagetto and A. Sciarrone, ``Vehicular/Non-Vehicular Multi-Class Multi-Object Tracking in Drone-Based Aerial Scenes,'' \emph{IEEE Transactions on Vehicular Technology}, vol. 73, no. 4, pp. 4961-4977, April 2024.

%\bibitem{Drone_ITS_TPMBM} Á. F. García-Fernández and J. Xiao, ``Trajectory Poisson Multi-Bernoulli Mixture Filter for Traffic Monitoring Using a Drone,'' \emph{IEEE Transactions on Vehicular Technology}, vol. 73, no. 1, pp. 402-413, Jan. 2024.

\bibitem{ITS_TPMBM} X. Zhang, H. Yu, Y. Qin, X. Zhou and S. Chan, ``Video-Based Multi-Camera Vehicle Tracking via Appearance-Parsing Spatio-Temporal Trajectory Matching Network,'' \emph{IEEE Transactions on Circuits and Systems for Video Technology}, vol. 34, no. 10, pp. 10077-10091, 2024.


\bibitem{GM-PHD_MOT} 
L. Lindenmaier, S. Aradi, T. Bécsi, O. Törő and P. Gáspár, ``GM-PHD Filter Based Sensor Data Fusion for Automotive Frontal Perception System," \emph{IEEE Transactions on Vehicular Technology}, vol. 71, no. 7, pp. 7215-7229, 2022.


% ** CAMO-MOT, IEEE T_ITS 2023 (Tested on both nuScenes and KITTI. NuScenes C+L 3D MOT 1st 20230112.)论文中Table I结果应该是用PointGNN作为3D Object detector的KITTI test dataset的front view 2D MOT result（当前公开论文的LiDAR and Camera fusion 3D MOT中在KITTI test dataset的front view 2D MOT result上性能最好，比PointRCNN + PC3T性能更牛。但CAMO-MOT方案的问题在于，代码没有开源，不确定它是online tracking的方案与否）。论文中Table III result是在validation dataset上的KITTI 3D MOT result
\bibitem{CAMO-MOT}
L. Wang, X. Zhang, W. Qin, X. Li, L. Yang, Z. Li, L. Zhu, H. Wang, and H. Liu, ``CAMO-MOT: Combined appearance-motion optimization for 3D multi-object tracking with camera-LiDAR fusion,'' \emph{IEEE Transactions on Intelligent Transportation Systems}, 2023, doi: 10.1109/TITS.2023.3285651.


% Our paper
%\bibitem{radar_instance_segmentation_1} J. Liu, W. Xiong, L. Bai, Y. Xia, T. Huang, W. Ouyang, and B. Zhu, ``Deep instance segmentation with automotive radar detection points,'' \emph{IEEE Transactions on Intelligent Vehicles}, vol. 8, no. 1, pp. 84-94, 2023.

%\bibitem{radar_instance_segmentation_2} W. Xiong, J. Liu, Y. Xia, T. Huang, B. Zhu, and W. Xiang, ``Contrastive learning for automotive mmWave radar detection points based instance segmentation,'' in \emph{Proceedings of the IEEE International Conference on Intelligent Transportation Systems (ITSC)}, 2022, pp. 1255-1261.

%\bibitem{radar_LiDAR_fusion_object_detection_RaLiBEV} Y. Yang, J. Liu, T. Huang, Q.-L. Han, G. Ma, and B. Zhu, ``RaLiBEV: Radar and LiDAR BEV fusion learning for anchor box free object detection systems," 2022, \emph{arXiv:2211.06108}. Submitted to IEEE Transactions on Intelligent Transportation Systems.

%\bibitem{SMURF} J. Liu, Q. Zhao, W. Xiong, T. Huang, Q.-L. Han, and B. Zhu, ``SMURF: Spatial multi-representation fusion for 3D object detection with 4D imaging radar," \emph{IEEE Transactions on Intelligent Vehicles}, Oct. 2023, doi: 10.1109/TIV.2023.3322729.

%\bibitem{LXL} W. Xiong, J. Liu, T. Huang, Q.-L. Han, Y. Xia, and B. Zhu, ``LXL: LiDAR excluded lean 3D object detection with 4D imaging radar and camera fusion,'' \emph{IEEE Transactions on Intelligent Vehicles}, Oct. 2023, doi: 10.1109/TIV.2023.3321240.





% 2D MOT
\bibitem{zhang2021fairmot}
Y. Zhang, C. Wang, X. Wang, W. Zeng, and W. Liu, ``FairMOT: 
On the fairness of detection and re-identification in multiple object tracking,'' \emph{International Journal of Computer Vision}, vol. 129, no. 11, pp. 3069-3087, Nov. 2021.

\bibitem{zhang2021bytetrack} 
Y. Zhang, P. Sun, Y. Jiang, D. Yu, Z. Yuan, P. Luo, W. Liu, and X. Wang, ``Bytetrack: Multi-object tracking by associating every detection box,'' in \emph{Proceedings of the European Conference on Computer Vision (ECCV)}, 2022, pp. 1-21.

\bibitem{Strongsort} 
Y. Du, Z. Zhao, Y. Song, Y. Zhao, F. Su, T. Gong, and H. Meng, ``StrongSort: Make deepsort great again,'' \emph{IEEE Transactions on Multimedia}, 2023, doi: 10.1109/TMM.2023.3240881.


% 3D Object Detector
\bibitem{PointRCNN}
S. Shi, X. Wang, and H. P. Li, ``3D object proposal generation and detection from point cloud,'' in \emph{Proceedings of the IEEE/CVF Conference on Computer Vision and Pattern Recognition (CVPR)}, 2019, pp. 16-20.

\bibitem{PointGNN}
W. Shi, and R. Rajkumar, ``Point-GNN: Graph neural network for 3D object detection in a point cloud,'' in \emph{Proceedings of the IEEE/CVF Conference on Computer Vision and Pattern Recognition (CVPR)}, 2020, pp. 1711-1719.

\bibitem{CenterPoint}
T. Yin, X. Zhou, and P. Krahenbuhl, ``Center-based 3D object detection and tracking, '' in \emph{Proceedings of the IEEE/CVF Conference on Computer Vision and Pattern Recognition (CVPR)}, 2021, pp. 11784-11793.

\bibitem{CasA}
H. Wu, J. Deng, C. Wen, X. Li, C. Wang, and J. Li, ``CasA: A cascade attention network for 3D object detection from LiDAR point clouds,'' \emph{IEEE Transactions on Geoscience and Remote Sensing}, vol. 60, pp. 1-11, 2022.





\bibitem{newcrf}
T. Gao, H. Pan, Z. Wang and H. Gao, "A CRF-based framework for tracklet inactivation in online multi-Object tracking," \emph{IEEE Transactions on Multimedia}, vol. 24, pp. 995-1007, 2022.

\bibitem{sqgrl}
W. Feng, L. Bai, Y. Yao, W. Gan, W. Wu and W. Ouyang, "Similarity- and quality-guided relation learning for joint detection and tracking," \emph{IEEE Transactions on Multimedia}, 2023, doi: 10.1109/TMM.2023.3279670.



\bibitem{Fantrack}
Baser, Erkan, Venkateshwaran Balasubramanian, Prarthana Bhattacharyya, and Krzysztof Czarnecki. "Fantrack: 3d multi-object tracking with feature association network." In 2019 IEEE Intelligent Vehicles Symposium (IV), pp. 1426-1433. IEEE, 2019.

\bibitem{mmMOT}
Zhang, Wenwei, Hui Zhou, Shuyang Sun, Zhe Wang, Jianping Shi, and Chen Change Loy. "Robust multi-modality multi-object tracking." In Proceedings of the IEEE/CVF international conference on computer vision, pp. 2365-2374. 2019.



% 3D MOT Early approach
\bibitem{AB3DMOT} 
X. Weng, J. Wang, D. Held, and K. Kitani, ``3D multi-object tracking: A baseline and new evaluation metrics,'' in \emph{Proceedings of the IEEE/RSJ International Conference on Intelligent Robots and Systems (IROS)}, 2020, pp. 10359-10366.
  
\bibitem{bayes}
S. Särkkä, and L. Svensson, ``Bayesian filtering and smoothing,`` \emph{Cambridge university press}, vol. 17, 2023.


%% LiDAR based 3D MOT using nuScenes dataset 

\bibitem{StanfordIPRL-TRI} 
H.-K. Chiu, A. Prioletti, J. Li, and J. Bohg, ``Probabilistic 3D multi-object tracking for autonomous driving,'' 2020, \emph{arXiv:2001.05673.}

% IRCA 2021
\bibitem{Pang2021PMBM} 
S. Pang, D. Morris, and H. Radha, ``3D multi-object tracking using random finite set-based multiple measurement models filtering (RFS-M3) for autonomous vehicles," in \emph{IEEE International Conference on Robotics and Automation (ICRA)}, 2021, pp. 13701-13707. 

% ICCV 2021
\bibitem{SimTrack} 
C. Luo, X. Yang, and A. Yuille, ``Exploring simple 3D multi-object tracking for autonomous driving,’’ in \emph{Proceedings of the IEEE/CVF International Conference on Computer Vision (ICCV)}, 2021, pp. 10488-10497.

% RAL 2022
\bibitem{OGR3MOT} 
J.-N. Zaech, A. Liniger, D. Dai, M. Danelljan, and L. Van Gool, ``Learnable online graph representations for 3D multi-object tracking,'' \emph{IEEE Robotics and Automation Letters}, vol. 7, no. 2, pp. 5103-5110, 2022.

% RAL 2022
\bibitem{Batch3DMOT}
M. Büchner, and A. Valada, ``3D multi-object tracking using graph neural networks with cross-edge modality attention,'' \emph{IEEE Robotics and Automation Letters}, vol. 7, no. 4, pp. 9707-9714, 2022.

\bibitem{SimpleTrack} 
Z. Pang, Z. Li, and N. Wang, ``SimpleTrack: Understanding and rethinking 3D multi-object tracking,'' in \emph{Proceedings of the European Conference on Computer Vision (ECCV) Workshop}, 2022, pp. 680–696.

\bibitem{ImmortalTracker} 
Q. Wang, Y. Chen, Z. Pang, N. Wang, and Z. Zhang, ``Immortal tracker: Tracklet never dies,'' 2021, \emph{arXiv:2111.13672.}

% Proceedings of the IEEE, 2018
\bibitem{BP_Tracker} 
F. Meyer, T. Kropfreiter, J. L. Williams, R. Lau, F. Hlawatsch, P. Braca, and M. Z. Win, ``Message passing algorithms for scalable multitarget tracking,'' in \emph{Proceedings of the IEEE}, vol. 106, no. 2, pp. 221-259, Feb. 2018.

% IV 2022
\bibitem{TransMOT} 
F. Ruppel, F. Faion, C. Gläser, and K. Dietmayer, ``Transformers for multi-object tracking on point clouds,'' in \emph{Proceedings of the IEEE Intelligent Vehicles Symposium (IV)}, 2022, pp. 832-838.

% ICRA 2022
\bibitem{PF-MOT} 
T. Wen, Y. Zhang, and N. M. Freris, ``PF-MOT: Probability fusion based 3D multi-object tracking for autonomous vehicles,'' in \emph{Proceedings of the International Conference on Robotics and Automation (ICRA)}, 2022, pp. 700-706.

\bibitem{ENBP_Tracker}
M. Liang and F. Meyer, "Neural enhanced belief propagation for multiobject tracking," \emph{IEEE Transactions on Signal Processing}, vol. 72, pp. 15-30, 2024.

% TIV 2023
\bibitem{GNN-PMB} J. Liu, L. Bai, Y. Xia, T. Huang, B. Zhu, and Q. -L. Han, ``GNN-PMB: A simple but effective online 3D multi-object tracker without bells and whistles," \emph{IEEE Transactions on Intelligent Vehicles}, vol. 8, no. 2, pp. 1176-1189, 2023.

\bibitem{Intertrack} 
J. Willes, C. Reading and S. Waslander, ``InterTrack: Interaction Transformer for 3D Multi-Object Tracking,'' in \emph{Proceedings of the Conference on Robots and Vision (CRV)}, 2023, pp. 73-80.

\bibitem{Minkowski_Tracker} 
J. Gwak, S. Savarese and J. Bohg, ``Minkowski tracker: A sparse spatio-temporal R-CNN for joint object detection and tracking,'' 2022, \emph{arXiv:2208.10056.}

\bibitem{ShaSTA} 
T. Sadjadpour, J. Li, R. Ambrus and J. Bohg, ``ShaSTA: Modeling Shape and Spatio-Temporal Affinities for 3D Multi-Object Tracking,'' \emph{IEEE Robotics and Automation Letters}, vol. 9, no. 5, pp. 4273-4280, May 2024.


\bibitem{bytetrackv2} 
Y. Zhang, X. Wang, X. Ye, W. Zhang, J. Lu, X. Tan, E. Ding, P. Sun, and J. Wang, ``ByteTrackV2: 2D and 3D multi-object tracking by associating every detection box,'' 2023, \emph{arXiv:2303.15334.} 



%% LiDAR based 3D MOT using KITTI dataset。注意KITTI官方榜单为KITTI test dataset的front view 2D MOT result。

% ** TIV 2023, 论文中Table I的result是在validation dataset上的KITTI 3D MOT result
\bibitem{ACK3DMOT}
G. Guo and S. Zhao, ``3D multi-object tracking with adaptive cubature Kalman filter for autonomous driving,'' \emph{IEEE Transactions on Intelligent Vehicles}, vol. 8, no. 1, pp. 84-94, 2023.

\bibitem{AdaptiveNoiseCov}
C. Jiang, Z. Wang, H. Liang and Y. Wang, ``A Novel Adaptive Noise Covariance Matrix Estimation and Filtering Method: Application to Multiobject Tracking,'' \emph{IEEE Transactions on Intelligent Vehicles}, vol. 9, no. 1, pp. 626-641, 2024.

% ** TMM 2023，论文中Table I的result是在validation dataset上的KITTI 3D MOT result。Table II的result是KITTI test dataset的front view 2D MOT result（效果较差）
\bibitem{CenterTube} 
H. Liu, Y. Ma, Q. Hu, and Y. Guo, ``CenterTube: Tracking multiple 3D objects with 4D tubelets in dynamic point clouds,'' \emph{IEEE Transactions on Multimedia}, vol. 25, pp. 8793-8804, 2023.

% ** ECCV 2022。KITTI官方榜单为KITTI test dataset的front view 2D MOT result（性能和EagerMOT相仿）
\bibitem{PolarMOT}
A. Kim, G. Brasó, A. Ošep, and L. Leal-Taixé, ``PolarMOT: How far can geometric relations take us in 3D multi-object tracking?,'' in \emph{Proceedings of the European Conference of Computer Vision (ECCV)}, 2022, pp. 41-58.

% ** KITTI官方榜单为CasA + UG3DMOT。论文中Table I result是KITTI test dataset的front view 2D MOT result（性能远不如PC3T，如果公平比较理论性能与EagerMOT相仿）
\bibitem{UG3DMOT}
J. He, C. Fu, and X. Wang, J. Wang, ``3D multi-object tracking based on informatic divergence-guided data association,'' \emph{Signal Processing}, vol. 222, pp. 109544, 2024.


% ** ICRA 2023（LiDAR only 3D MOT中在KITTI test dataset front view 2D MOT result上比PointRCNN + PC3T性能还好，但论文暂时没有公开。根据KITTI官方描述该方案大概也是一个off smoother: https://www.cvlibs.net/datasets/KITTI/eval_tracking_detail.php?result=76b4d264326d96680e7311363960fe1b8b933c5f）
\bibitem{Rethinking3DMOT}
L. Wang, J. Zhang, P. Cai, and X. Li, ``Towards robust reference system for autonomous driving: Rethinking 3D MOT,'' in \emph{Proceedings of the IEEE International Conference on Robotics and Automation (ICRA)}, 2023, pp. 8319-8325.

% ** TITS 2022, KITTI官方榜单有两个result: PC3T, CasTrack。前者是PC3T tracker + 某种比CasA能力弱的detector（应该是PointRCNN）的KITTI test dataset front view 2D MOT result，后者是PC3T tracker + CasA的KITTI test dataset front view 2D MOT result（当前公开论文的LiDAR only 3D MOT中在KITTI test dataset front view 2D MOT result上性能次好，然而其在KITTI榜单上较优秀的性能指标并不是online tracking的结果，而是加入了post-processing的offline smoothing的output，猜测使用online tracking版本的PC3T性能也与EagerMOT接近）
\bibitem{PC3T} 
H. Wu, W. Han, C. Wen, X. Li, and C. Wang, ``3D multi-object tracking in point clouds based on prediction confidence-guided data association,'' \emph{IEEE Transactions on Intelligent Transportation Systems}, vol. 23, no. 6, pp. 5668-5677, 2022.

% ** IJCAI 2021. 本文给的KITTI test dataset front view 2D MOT result（当前公开论文的LiDAR only 3D MOT中在KITTI test dataset front view 2D MOT result上性能最好，然而本质上也是一个offline smoothing）。
\bibitem{PC-TCNN}
H. Wu, Q. Li, C. Wen, X. Li, X. Fan, and C. Wang, ``Tracklet proposal network for multi-object tracking on point clouds,'' in \emph{Proceedings of the International Joint Conferences on Artificial Intelligence (IJCAI)}, 2021, pp. 1165--1171.





%% Camera and LiDAR Fusion based 3D MOT using nuScenes dataset

% ICRA 2021
\bibitem{Probabilistic3DMM} 
H.-K. Chiu, J. Li, R. Ambruş, and J. Bohg, ``Probabilistic 3D multi-modal, multi-object tracking for autonomous driving,'' in \emph{Proceedings of the IEEE International Conference on Robotics and Automation (ICRA)}, 2021, pp. 14227-14233.

% IROS 2021
\bibitem{CBMOT} 
N. Benbarka, J. Schröder, and A. Zell, ``Score refinement for confidence-based 3D multi-object tracking,'' in \emph{Proceedings of IEEE/RSJ International Conference on Intelligent Robots and Systems (IROS)}, 2021, pp. 8083-8090.

\bibitem{MF-Net} 
S. Tian, M. Duan, J. Deng, H. Luo and Y. Hu, ``MF-Net: A Multimodal Fusion Model for Fast Multi-object Tracking,'' \emph{IEEE Transactions on Vehicular Technology}, 2024, doi: 10.1109/TVT.2024.3375457.



% IROS 2021
\bibitem{AlphaTrack} 
Y. Zeng, C. Ma, M. Zhu, Z. Fan, and X. Yang, ``Cross-modal 3D object detection and tracking for auto-driving,'' in \emph{Proceedings of the 2021 IEEE/RSJ International Conference on Intelligent Robots and Systems (IROS)}, 2021, pp. 3850-3857.


%\bibitem{newtt} S. Scheidegger, J. Benjaminsson, E. Rosenberg, A. Krishnan, and K. Granström, ``Mono-camera 3D multi-object tracking using deep learning detections and PMBM filtering,'' in \emph{Proceedings of the IEEE Intelligent Vehicles Symposium (IV)}, June 2018, pp. 433-440.

%% Camera and LiDAR Fusion based 3D MOT using KITTI dataset。注意KITTI官方榜单为KITTI test dataset的front view 2D MOT result。

\bibitem{GNNMOT}
Li, Jiahe, Xu Gao, and Tingting Jiang. "Graph networks for multiple object tracking." In Proceedings of the \emph{IEEE/CVF winter conference on applications of computer vision}, pp. 719-728. 2020.

\bibitem{Gnn3dmot}
X. Weng, Y. Wang, Y. Man, and K. M. Kitani, ``GNN3DMOT: Graph neural network for 3D multi-object tracking with 2D-3D multi-feature learning,'' in Proceedings of the \emph{IEEE Conference on Computer Vision and Pattern Recognition (CVPR)}, 2020, pp. 6499–6508.

\bibitem{MOTG}
Dai, Peng, Renliang Weng, Wongun Choi, Changshui Zhang, Zhangping He, and Wei Ding. "Learning a proposal classifier for multiple object tracking." In Proceedings of the \emph{IEEE/CVF Conference on Computer Vision and Pattern Recognition (CVPR)}, pp. 2443-2452. 2021.

\bibitem{MOTG2}
He, Jiawei, Zehao Huang, Naiyan Wang, and Zhaoxiang Zhang. "Learnable graph matching: Incorporating graph partitioning with deep feature learning for multiple object tracking." In Proceedings of the \emph{IEEE/CVF conference on computer vision and pattern recognition (CVPR)}, pp. 5299-5309. 2021.


\bibitem{ConvUKF}
S. Liu, W. Cao, C. Liu, T. Zhang and S. E. Li, ``Convolutional Unscented Kalman Filter for Multi-Object Tracking With Outliers,'' \emph{IEEE Transactions on Intelligent Vehicles}, 2024, doi: 10.1109/TIV.2024.3446851.

\bibitem{Real_Time_MOT}
S. Feng, X. Li, Z. Yan, S. Li, Y. Zhou, C. Xia, and X. Wang, ``Accurate and Real-Time 3D-LiDAR Multi-Object Tracking Using Factor Graph Optimization,'' \emph{IEEE Sensors Journal}, vol. 24, no. 2, pp. 1760-1771, 2024.

% ** IEEE TITS 2023。
\bibitem{IMSF-MOT}
G. Wang, C. Peng, Y. Gu, J. Zhang, and H. Wang, ``Interactive multi-scale fusion of 2D and 3D features for multi-object vehicle tracking,'' \emph{IEEE Transactions on Intelligent Transportation Systems},  vol. 24, no. 10, pp. 10618-10627, Oct 2023.

% ** ICRA 2021，论文中Table IIresult是在validation dataset上的KITTI 3D MOT result。Table III结果应该是用PointGNN（CVPR2020的一种比CVPR 2019年Point RCNN更晚一点的目标检测器）作为3D Object Detector（同时RRC作为2D Object detector）的KITTI test dataset的front view 2D MOT result。
\bibitem{EagerMOT} 
A. Kim, A. Ošep, and L. Leal-Taixé, ``EagerMOT: 3D multi-object tracking via sensor fusion,'' in \emph{Proceedings of the IEEE International Conference on Robotics and Automation (ICRA)}, 2021, pp. 11315-11321.

\bibitem{JMODT}
K. Huang, and Q. Hao, ``Joint multi-object detection and tracking with camera-LiDAR fusion for autonomous driving,'' in \emph{Proceedings of the IEEE/RSJ International Conference on Intelligent Robots and Systems (IROS)}, 2021, pp. 6983-6989.

% ** TIV 2022，论文中Table I结果是用PointGNN（CVPR2020的一种比CVPR 2019年Point RCNN更晚一点的目标检测器）作为3D Object Detector（同时RRC作为2D Object detector）的KITTI test dataset的front view 2D MOT result。（性能和EagerMOT相仿）
\bibitem{DualTracker}
Y. Ma, J. Zhang, G. Qin, J. Jin, K. Zhang, D. Pan, and M. Chen, ``3D multi-object tracking based on dual-tracker and DS evidence theory,'' \emph{IEEE Transactions on Intelligent Vehicles}, vol. 8, no. 3, pp. 2426-2436, 2023.

% ** RAL 2022，论文中Table II结果应该是用PointRCNN作为3D Object detector的KITTI test dataset的front view 2D MOT result（性能和EagerMOT相仿）。
\bibitem{DeepFusionMOT}
X. Wang, C. Fu, Z. Li, Y. Lai and J. He, ``DeepFusionMOT: A 3D multi-object tracking framework based on camera-LiDAR fusion with deep association,'' \emph{IEEE Robotics and Automation Letters}, vol. 7, no. 3, pp. 8260-8267, 2022.

% ** IEEE Sensors Journal 2022，论文中Table I结果应该是用PointRCNN作为3D Object detector的KITTI test dataset的front view 2D MOT result（性能和EagerMOT相仿）。
\bibitem{StrongFusionMOT}
X. Wang, C. Fu, J. He, S. Wang, and J. Wang, ``StrongFusionMOT: A multi-object tracking method based on LiDAR-camera fusion,'' \emph{IEEE Sensors Journal}, Dec 2022, doi: 10.1109/JSEN.2022.3226490.

\bibitem{Feng}
S. Feng, X. Li, Z. Yan, C. Xia, S. Li, X. Wang, and Y. Zhou, ``Tightly Coupled Integration of LiDAR and Vision for 3D Multiobject Tracking,'' \emph{IEEE Transactions on Intelligent Vehicles}, 2024, doi: 10.1109/TIV.2024.3413733.

% ** IEEE TVT 2023，论文中Table VIII结果应该是用Point-GNN作为3D Object detector的KITTI test dataset的front view 2D MOT result（性能和StrongFusionMOT相仿）。
\bibitem{HIDMOT}
Y. An, J. Wu, Y. Cui, and H. Hu, ``Multi-object tracking based on a novel feature image with multi-modal information,'' \emph{IEEE Transactions on Vehicular Technology}, vol. 72, no. 8, pp. 9909-9921, Aug. 2023.

% Sensor 2022
\bibitem{MSA-MOT}
Z. Zhu, J. Nie, H. Wu, Z. He, and M. Gao, ``MSA-MOT: Multi-stage association for 3D multimodality multi-object tracking,'' \emph{Sensors}, vol. 22, no. 22, pp. 8650, 2022.


\bibitem{AdverseWeather-MOT}
L. Qiao, P. Zhang, Y. Liang, X. Yan, L. Huangfu, X. Zheng, and Z. Yu, ``Cross-Modality 3D Multi-Object Tracking Under Adverse Weather via Adaptive Hard Sample Mining,'' \emph{IEEE Internet of Things Journal}, vol. 11, no. 14, pp. 25268-25282, July, 2024.

\bibitem{ptp}
Weng, X., Yuan, Y. and Kitani, K., `` PTP: Parallelized tracking and prediction with graph neural networks and diversity sampling,'' \emph{IEEE Robotics and Automation Letters}, pp.4640-4647, 2021.

\bibitem{3dmotformer}
Ding, S., Rehder, E., Schneider, L., Cordts, M. and Gall, J. ``3dmotformer: Graph transformer for online 3d multi-object tracking,'' in \emph{Proceedings of the IEEE/CVF International Conference on Computer Vision}, 2023, pp. 9784-9794.



% ** DFR-FastMOT, Arxiv 2023。论文中Table II结果应该是用PointRCNN作为3D Object detector, RCC作为3D object detector在KITTI validation dataset的front view 2D MOT result（性能非常好，HOTA大概可以到达84.28%）。
%\bibitem{DFR-FastMOT}
%M. Nagy, M. Khonji, J. Dias, and S. Javed, ``DFR-FastMOT: Detection Failure Resistant Tracker for Fast Multi-Object Tracking Based on Sensor Fusion,'' 2023, \emph{arXiv:2302.14807}.


% CasA 3D detector, RCNN 2D detector
\bibitem{MMF-JDT}
X. Wang et al., "A Multi-Modal Fusion-Based 3D Multi-Object Tracking Framework with Joint Detection," \emph{IEEE Robotics and Automation Letters}, vol. 10, no. 1, pp. 532-539, 2025.








\bibitem{MTonlnea}
C. H. Kuo, C. Huang, and R. Nevatia `` Multi-target tracking by online learned discriminative appearance models``, \emph{IEEE Conference on Computer Vision and Pattern Recognition (CVPR)}, 2010, pp. 685-692.

\bibitem{lstm}
A. Alahi, K. Goel, V. Ramanathan, A. Robicquet, F.-F. Li, and S. Savarese, ''Social LSTM: Human trajectory prediction in crowded spaces,''  in \emph{Proceedings of the IEEE Conference on Computer Vision and Pattern Recognition}, 2016, pp. 961-971.

\bibitem{hu2019joint}
H. N. Hu, Q.Z. Cai, D. Wang, J. Lin, M. Sun, P. Krahenbuhl, T. Darrell, and F. Yu, ``Joint monocular 3D vehicle detection and tracking,`` \emph{IEEE/CVF International Conference on Computer Vision}, 2019, pp. 5390-5399.

\bibitem{qi2017pointnet}
C. R. Qi, H. Su, K. Mo, and L.J. Guibas, ``Pointnet: Deep learning on point sets for 3D classification and segmentation.`` \emph{IEEE Conference on Computer Vision and Pattern Recognition}, 2017, pp. 652-660.

\bibitem{jiang2019graph}
X. Jiang, P. Li, Y. Li, and X. Zhen, `` Graph neural based end-to-end data association framework for online multiple-object tracking,`` 2019, \emph{arXiv:1907.05315}.

\bibitem{velivckovic2017graph}
P. Veličković, G. Cucurull, A. Casanova, A. Romero, P. Lio and Y. Bengio, ``Graph attention networks,`` in \emph{Proceedings of the International Conference on Learning Representations (ICLR)}, 2018.

% \bibitem{fang2018recurrent}
% Fang, K., Xiang, Y., Li, X. and Savarese, S. ``Recurrent autoregressive networks for online multi-object tracking.`` \emph{IEEE Winter Conference on Applications of Computer Vision (WACV)}, March 2018, pp. 466-475, doi: 10.1109/WACV.2018.00057.


\bibitem{blackman1999design} 
B. Samuel, and R. Popoli. ''Design and analysis of modern tracking systems,'' \emph{Artech House}, 1999. 


% \bibitem{chui2017kalman}
% Chui, C.K. and Chen, G., 2017. ``Kalman filtering,`` \emph{Berlin, Germany: Springer International Publishing}, 2017, pp. 19-26.







% % JPDA Theory 1
% \bibitem{JPDA_Theory_1} 
% T. Fortmann, Y. Bar-Shalom, and M. Scheffe, ``Sonar tracking of multiple targets using joint probabilistic data association,'' \emph{IEEE Journal of Oceanic Engineering}, vol. 8, no. 3, pp. 173-184, July 1983.

% % JPDA Theory 2
% \bibitem{BJPDA_Theory_2} 
% Y. Bar-Shalom, F. Daum, and J. Huang, ``The probabilistic data association filter,'' \emph{IEEE Control Systems Magazine}, vol. 29, no. 6, pp. 82-100, Dec. 2009.

% % MHT Theory 1
% \bibitem{MHT_Theory} 
% D. Reid, ``An algorithm for tracking multiple targets,'' \emph{IEEE Transactions on Automatic Control}, vol. 24, no. 6, pp. 843-854, Dec. 1979.

% % MHT Theory 2
% \bibitem{Blackman2004MultipleHT}  
% S. S. Blackman, ``Multiple hypothesis tracking for multiple target tracking,'' \emph{IEEE Aerospace and Electronic Systems Magazine}, vol. 19, no. 1, pp. 5-18, Jan. 2004.
\bibitem{hota}
Luiten, J., A. Osep, P. Dendorfer, P. Torr, A. Geiger, L. Leal-Taixé, and B. Leibe Hota. "A higher order metric for evaluating multi-object tracking., 2021, 129." DOI: https://doi. org/10.1007/s11263-020-01375-2. PMID: https://www. ncbi. nlm. nih. gov/pubmed/33642696: 548-578.

% Dataset KITTI
\bibitem{Geiger2013VisionMR} 
A. Geiger, L. Philip, S. Christoph, and U. Raquel,  ``Vision meets robotics: The KITTI dataset,'' \emph{The International Journal of Robotics Research}, vol. 32, no. 11, pp. 1231-1237, Aug. 2013.

% Dataset Waymo
\bibitem{Waymodata}
P. Sun, H. Kretzschmar, X. Dotiwalla, A. Chouard, V. Patnaik, P. Tsui, J. Guo, Y. Zhou, Y. Chai, and B. Caine, ``Scalability in Perception for Autonomous Driving: Waymo Open Dataset. '' \emph{IEEE Conference on Computer Vision and Pattern Recognition (CVPR)}, 2020.

% % Dataset Waymo
\bibitem{Waymo} 
P. Sun, H. Kretzschmar, X. Dotiwalla, A. Chouard, V. Patnaik, P. Tsui, J. Guo, Y. Zhou, Y. Chai, B. Caine, and V. Vasudevan, ``Scalability in perception for autonomous driving: Waymo open dataset,'' in \emph{Proceedings of the IEEE/CVF Conference on Computer Vision and Pattern Recognition (CVPR)}, 2020, pp. 2446-2454.

\bibitem{3DMODT}
J. Kini, A. Mian and M. Shah, ``3DMODT: Attention-Guided Affinities for Joint Detection and Tracking in 3D Point Clouds,'' in \emph{Proceedings of the IEEE International Conference on Robotics and Automation (ICRA)}, 2023, pp. 841-848.

%RAL 2024
\bibitem{Fast-Poly}
X. Li, D. Liu, L. Zhao, Y. Wu, X. Wu, and J. Gao, ``Fast-Poly: A Fast Polyhedral Framework For 3D Multi-Object Tracking,'' \emph{IEEE Robotics and Automation Letters}, vol. 9, no. 11, pp. 10519-10526, 2024.





% % Dataset Argoverse
% \bibitem{Chang2019Argoverse3T} 
% M.-F. Chang et al., ``Argoverse: 3D tracking and forecasting with rich maps,'' in \emph{Proceedings of the IEEE/CVF Conference on Computer Vision and Pattern Recognition (CVPR)}, 2019, pp. 8740-8749.

% \bibitem{Nuscenes} 
% H. Caesar et al., “nuScenes: A. multimodal dataset for autonomous driving" in \emph{Proceedings of the IEEE/CVF Conference on Computer Vision and Pattern Recognition (CVPR)}, 2020, pp. 11621–11631.

\end{thebibliography}

%\iffalse
% biography section
\vspace{-10mm}
\begin{IEEEbiography}[{\includegraphics[width=1in,height=1.25in,clip,keepaspectratio]{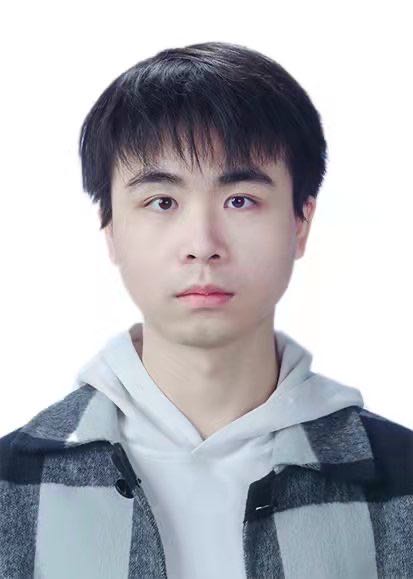}}]{Zhenrong Zhang} received his M.Sc. degree in Advanced Computer Science from the University of Manchester in 2020. Since 2022, he has been a Ph.D. student in the School of AI and Advanced Computing at Xi'an Jiaotong-Liverpool University, Suzhou. His primary research interests lie in the fields of computer vision and multi-object tracking.
\end{IEEEbiography}

\vspace{-10mm}
\begin{IEEEbiography}[{\includegraphics[width=1in,height=1.25in,clip,keepaspectratio]{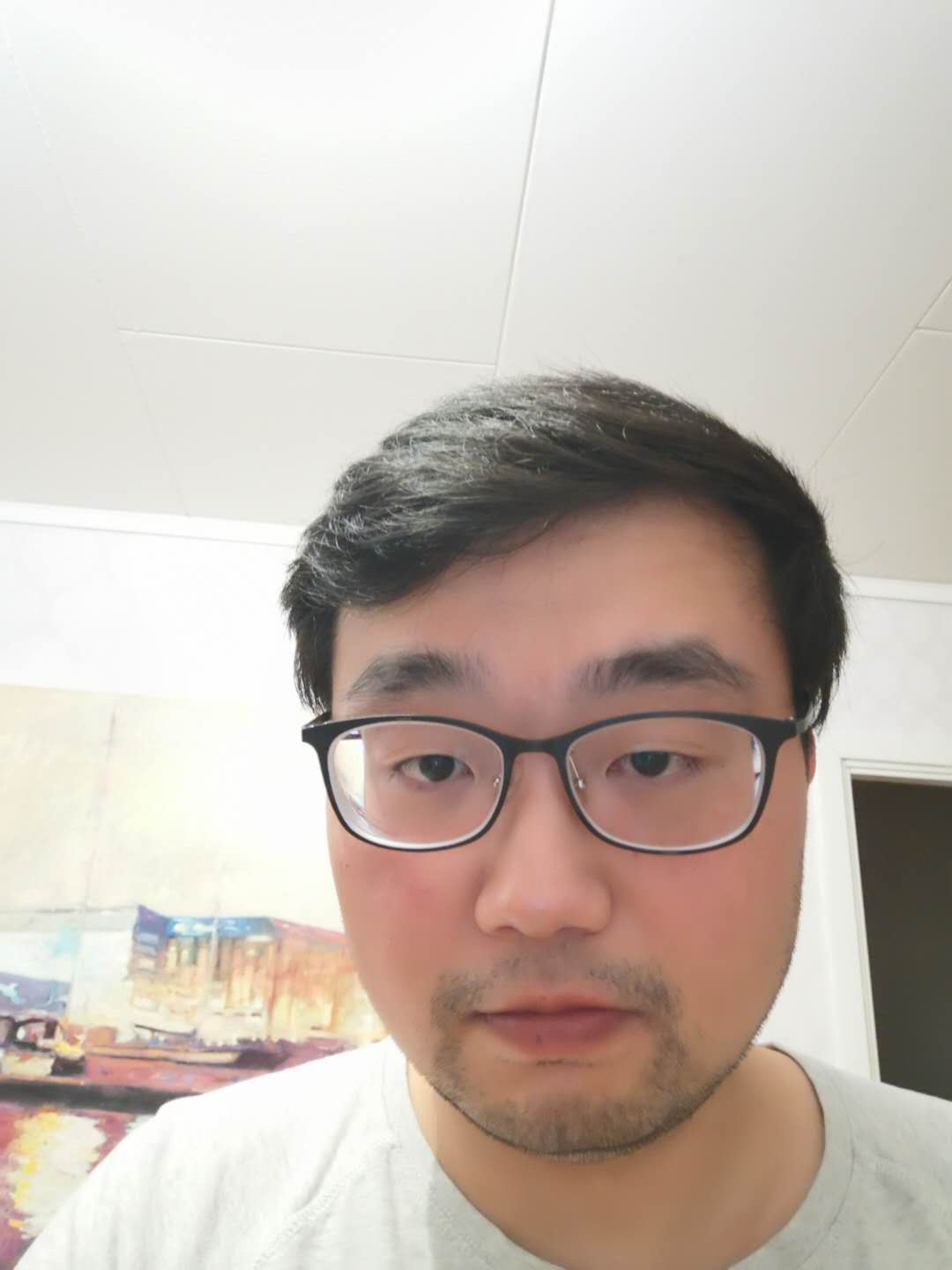}}]{Jianan Liu} received his B.Eng. degree in Electronics and Information Engineering from Huazhong University of Science and Technology, Wuhan, China, in 2007. He received his M.Eng. degree in Telecommunication Engineering from the University of Melbourne, Australia, and his M.Sc. degree in Communication Systems from Lund University, Sweden, in 2009 and 2012, respectively.
Jianan has over ten years of experience in software and algorithm design and development. He has held senior R\&D roles in the AI consulting, automotive, and telecommunication industries.
His research interests include applying statistical signal processing and deep learning for medical image processing, wireless communications, IoT networks, indoor sensing, and outdoor perception using a variety of sensor modalities like radar, camera, LiDAR, WiFi, etc.
\end{IEEEbiography}

\vspace{-10mm}
\begin{IEEEbiography}[{\includegraphics[width=1in,height=1.25in,clip,keepaspectratio]{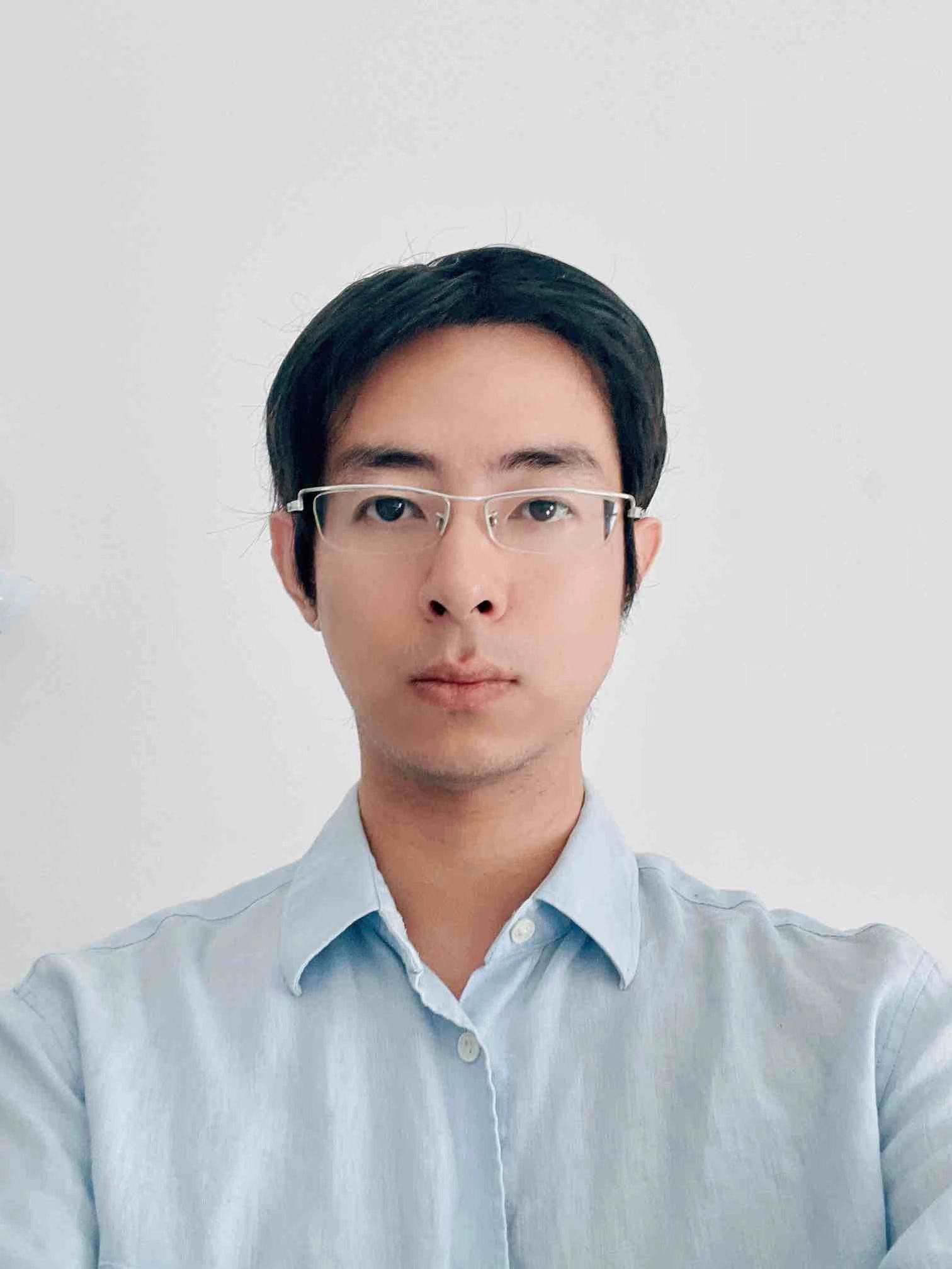}}]{Yuxuan Xia} (Member, IEEE) received his M.Sc. in communication engineering and Ph.D. in signal and systems from Chalmers University of Technology, Gothenburg, Sweden, in 2017 and 2022, respectively. After obtaining his Ph.D., he first stayed at the Signal Processing group, Chalmers University of Technology as a postdoctoral researcher for a year, and then he was with Zenseact AB and the Division of Automatic Control, Linkoping University as an Industrial Postdoctoral researcher for a year. He is currently a researcher at the Department of Automation, Shanghai Jiaotong University. His main research interests include sensor fusion, multi-object tracking and SLAM, especially for automotive applications. He has organized tutorials on multiobject tracking at the 2020-2024 FUSION conferences and the 2024 MFI conference. He has received paper awards at 2021 FUSION and 2024 MFI.
\end{IEEEbiography}

\vspace{-10mm}
\begin{IEEEbiography}[{\includegraphics[width=1in,height=1.25in,clip,keepaspectratio]{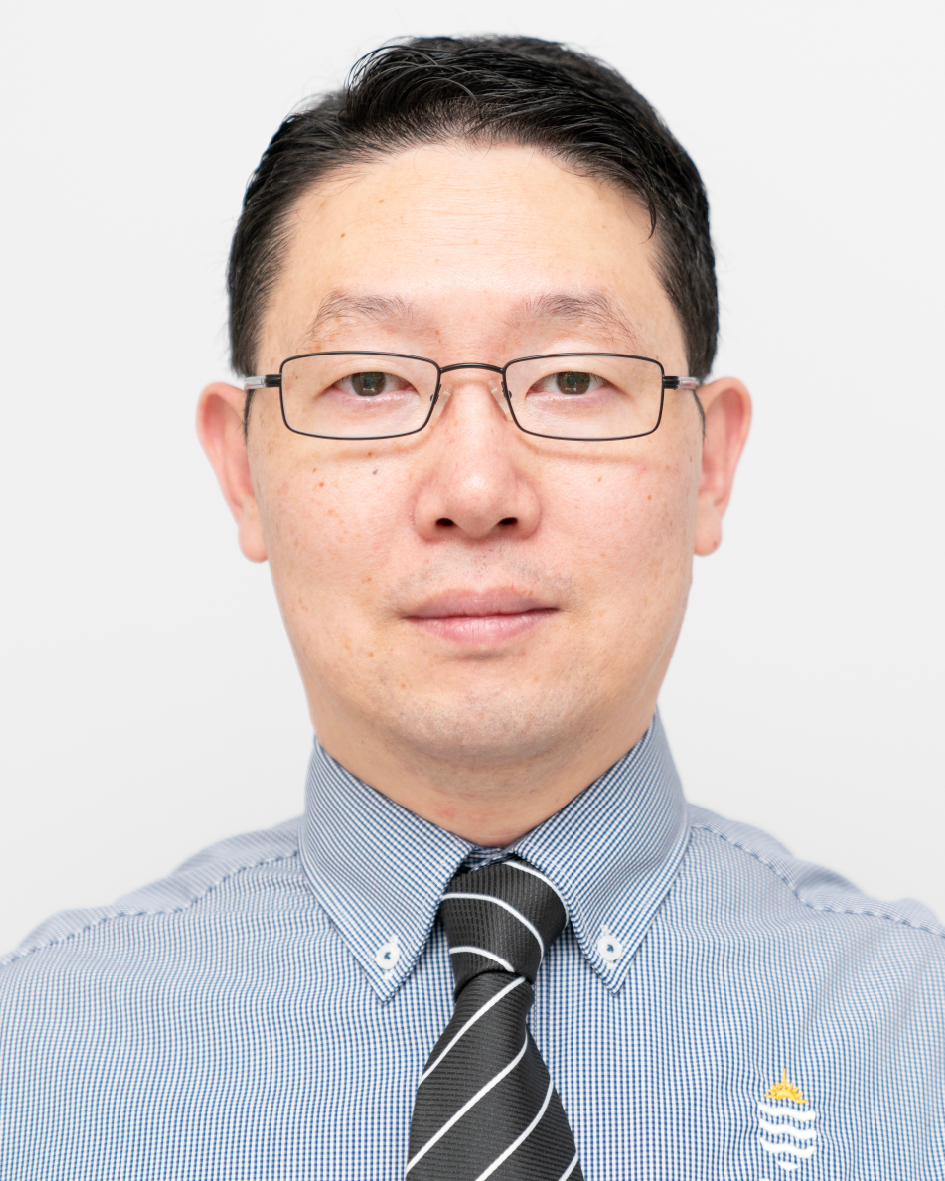}}]{Tao Huang} (Senior Member, IEEE) received the Ph.D. degree in Electrical Engineering from the University of New South Wales, Australia, in 2016. He received the M.Eng. degree in Sensor System Signal Processing from the University of Adelaide, Australia, in 2007, and the B.Eng. degree in Electronics and Information Engineering from Huazhong University of Science and Technology, China, in 2003. He is currently a Senior Lecturer and College Head of International Partnerships at James Cook University (JCU), Cairns, Australia. Dr. Huang was an Endeavour Australia Cheung Kong Research Fellow, a visiting scholar at the Chinese University of Hong Kong, a research associate at the University of New South Wales, and a postdoctoral research fellow at JCU. Prior to joining academia, he held industry roles as a senior engineer, senior data scientist, project team lead, and technical lead. He has authored or coauthored over 90 publications, including journal articles, conference papers, book chapters, and edited volumes. He is also a co-inventor of an international patent on MIMO systems. His awards include the Australian Postgraduate Award, the Engineering Research Award from the University of New South Wales, the Best Paper Award at IEEE WCNC (2011), the IEEE Outstanding Leadership Award (2022), IEEE Access Outstanding Associate Editor (2023 and 2024), and the JCU Citation for Outstanding Contribution to Student Learning (2022). Dr. Huang currently serves as Vice Chair of the IEEE Northern Australia Section and Chair of the local MTT-S/ComSoc Chapter. He previously served as Chair of the IEEE Young Professionals Affinity Group for the same section. He has also held leadership roles in international conferences, including workshop co-chair, publication co-chair, technical program committee chair, program vice chair, and symposium chair. He is an Associate Editor for the IEEE Open Journal of the Communications Society, IEEE Access, and IET Communications. His research interests include deep learning, intelligent sensing, computer vision, pattern recognition, wireless communications, system optimization, electronic systems, and IoT security.
 
\end{IEEEbiography}

\vspace{-10mm}
\begin{IEEEbiography}[{\includegraphics*[width=1in,height=1.25in,clip,keepaspectratio]{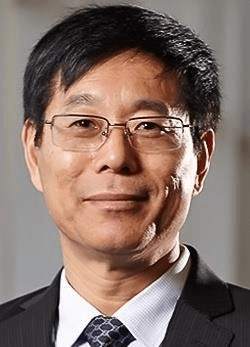}}]{Qing-Long Han} (Fellow, IEEE) received the B.Sc. degree in Mathematics 
from Shandong Normal University, Jinan, China, in 1983, and the M.Sc. and Ph.D. degrees in Control Engineering from East China University of Science and Technology, Shanghai, China, in 1992 and 1997, respectively.

Professor Han is Pro Vice-Chancellor (Research Quality) and a Distinguished Professor at Swinburne University of Technology, Melbourne, Australia. He held various academic and management positions at Griffith University and Central Queensland University, Australia. His research interests include networked control systems, multi-agent systems, time-delay systems, smart grids, unmanned surface vehicles, and neural networks.

Professor Han was awarded the 2024 IEEE Dr.-Ing. Eugene Mittelmann Achievement Award (the Highest Award in Industrial Electronics),  the 2024 Chinese Association of Automation (CAA) Science and Technology Achievement Award (the Highest Achievement Award of CAA in Automation, Information and Intelligent Science), the 2021 Norbert Wiener Award (the Highest Award in Systems Science and Engineering, and Cybernetics), and the 2021 M. A. Sargent Medal (the Highest Award of the Electrical College Board of Engineers Australia). He was the recipient of the IEEE Systems, Man, and Cybernetics Society Andrew P. Sage Best Transactions Paper Award in 2019, 2020, and 2022, respectively, the IEEE/CAA Journal of Automatica Sinica Norbert Wiener Review Award in 2020, and the IEEE Transactions on Industrial Informatics Outstanding Paper Award in 2020. 

Professor Han is a Member of the Academia Europaea (The Academy of Europe). He is a Fellow of the International Federation of Automatic Control (FIFAC), an Honorary Fellow of the Institution of Engineers Australia (HonFIEAust), and a Fellow of the Chinese Association of Automation (FCAA). He is a Highly Cited Researcher in both Engineering and Computer Science (Clarivate). He has served as an AdCom Member of IEEE Industrial Electronics Society (IES), a Member of IEEE IES Fellows Committee, a Member of IEEE IES Publications Committee,  Chair of IEEE IES Technical Committee on Network-Based Control Systems and Applications, and the Co-Editor-in-Chief of IEEE Transactions on Industrial Informatics. He is currently the President-Elect, an Executive Board Member, and a Steering Committee Member of Asian Control Association (ACA). He is currently the Editor-in-Chief of IEEE/CAA Journal of Automatica Sinica.

\end{IEEEbiography}

\vspace{-10mm}
\begin{IEEEbiography}[{\includegraphics[width=1in,height=1.25in,clip,keepaspectratio]{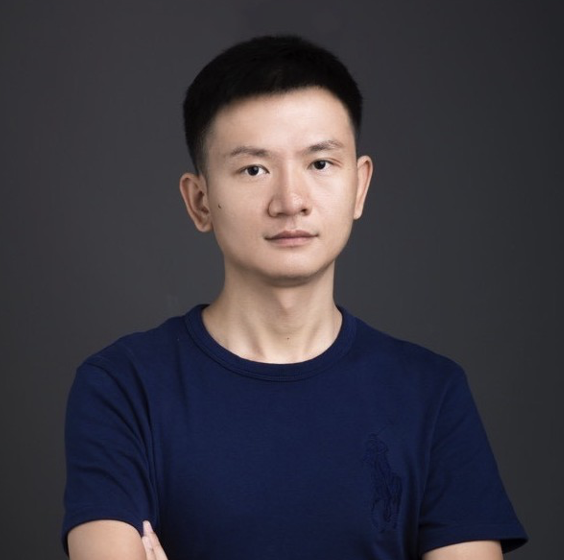}}]{Hongbin Liu}received his M.Sc. of Computer Science from RMIT University, Australia, in 2012, and his Ph.D degree from James Cook University, Australia, in 2020 respectively.  He is currently an assistant professor in School of Artificial Intelligence and Advanced Computing in Xi’an Jiaotong-Liverpool University, Suzhou, China. Prior to his PhD, he accumulated over five years of industry experience as a software developer, senior software developer, and project lead. Before joining Xi'an Jiaotong-Liverpool University, Hongbin served as a sessional lecturer at James Cook University for more than a year, where he taught various data science subjects. Hongbin's research interests revolve around Artificial Intelligence, with a particular focus on Spatio-Temporal Reasoning and Computer Vision.
\end{IEEEbiography}
%\fi

\end{document}